\documentclass[journal]{IEEEtran}
\usepackage{bm}
\ifCLASSINFOpdf
  \usepackage[pdftex]{graphicx}
\else
\fi
\usepackage{amsmath}
\usepackage{amssymb}  
\usepackage{mathptmx} 

\usepackage{algorithmic}

\usepackage{array}

\ifCLASSOPTIONcompsoc
 \usepackage[caption=false,font=normalsize,labelfont=sf,textfont=sf]{subfig}
\else
 \usepackage[caption=false,font=footnotesize]{subfig}
\fi

\usepackage{multirow}

\begin{document}
%
\title{Time-Optimal Handover Trajectory Planning for Aerial Manipulators based on Discrete Mechanics and Complementarity Constraints}
%
%
%

\author{Wei Luo, Jingshan Chen, Henrik Ebel, Peter Eberhard*~\thanks{The authors are with Institute of Engineering and Computational Me- chanics, University of Stuttgart, Pfaffenwaldring 9, 70569 Stuttgart, Germany \{wei.luo, jingshan.chen, henrik.ebel, peter.eberhard\}@itm.uni-stuttgart.de.}}

\IEEEspecialpapernotice{(Preprint)}

\maketitle

\begin{abstract}
  Planning a time-optimal trajectory for aerial robots is critical in many drone applications, such as rescue missions and package delivery, which have been widely researched in recent years. However, it still involves several challenges, particularly when it comes to incorporating special task requirements into the planning as well as the aerial robot's dynamics. In this work, we study a case where an aerial manipulator shall hand over a parcel from a moving mobile robot in a time-optimal manner. Rather than setting up the approach trajectory manually, which makes it difficult to determine the optimal total travel time to accomplish the desired task within dynamic limits, we propose an optimization framework, which combines discrete mechanics and complementarity constraints (DMCC) together. In the proposed framework, the system dynamics is constrained with the discrete variational Lagrangian mechanics that provides reliable estimation results also according to our experiments. The handover opportunities are automatically determined and arranged based on the desired complementarity constraints. Finally, the performance of the proposed framework is verified with numerical simulations and hardware experiments with our self-designed aerial manipulators.
\end{abstract}

\begin{IEEEkeywords}
  trajectory planning, aerial manipulator, discrete mechanics and complementarity constraints (DMCC), discrete Lagrangian mechanics, dynamic handover.
\end{IEEEkeywords}

\IEEEpeerreviewmaketitle

\section{Introduction}
\IEEEPARstart{F}{lying} unmanned aerial vehicles (UAVs), especially multi-rotors, have gained increasing attention over recent years. Compared to other aerial robots, multi-rotors are light-weight and more flexible, and they can hover over the ground and be operated in a small space. With various sensors onboard, such as LIDAR-sensors~\cite{ChiellaTeixeiraPereira2019} and cameras~\cite{AitJellalZell2017}, multi-rotors can be used for photographic and supervisory control duties~\cite{BorkarChowdhary2021}. Additionally, one can equip multi-rotors with additional actors or manipulators to create aerial manipulators, which can further enhance and extend the multi-rotors' functional capabilities, so that they are capable of transporting parcels in the 3D space~\cite{LoiannoKumar2018} and even doing delicate tasks at a high altitude~\cite{DingLuWangDing2021}.

However, there are still several challenges for applying multi-rotors. On the one hand, the onboard battery capacity is limited, so that their operation time and the available payload are restricted, and finding an energy-efficient trajectory is demanded for most drone applications. While the energy-efficient trajectory is not always a time-optimal trajectory, overall the energy consumption can indeed correlate to the total travel time~\cite{AlkomyShan2022}. Therefore, the shortest possible travel time is usually preferred, even more so in the case of drone racing~\cite{SpicaCristofaloWangMontijanoSchwager2020,DelmericoCieslewskiRebecqFaesslerScaramuzza2019} and timely package delivery~\cite{YuanRodrigues2019,KloetzerBurlacuEnescuCaraimanMahulea2019}. On the other hand, adding sensors or mechanical components to the multi-rotor system increases its mechanical complexity, complicating both control of the multi-rotor and planning an admissible trajectory, particularly when one wants to drive the system to its dynamic limit without violating the dynamical constraints.

\begin{figure}[htpb]
  \centering
  \includegraphics[scale=1]{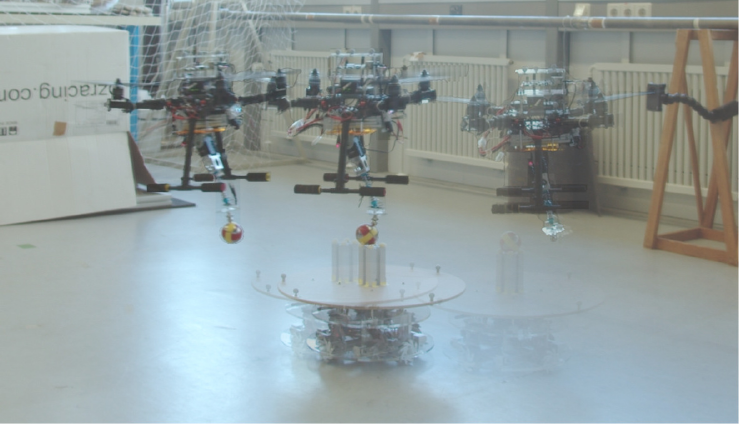}
  \caption{A heterogeneous robot system composed of an aerial manipulator and a ground mobile robot}
  \label{fig:investigated_scenario}
\end{figure}

In this work, we conduct an investigation of a heterogeneous robotic system in which a flying quadrotor equipped with an onboard one-degree of freedom (1-DoF) manipulator wants to approach and hand over an object with a moving mobile robot, as illustrated in Fig.~\ref{fig:investigated_scenario}. For a flexible and efficient object handover, rather than requiring the mobile robot to halt and wait for the aerial manipulator, this work aims to determine an optimal trajectory for the quadrotor and its onboard manipulator while the mobile robot can continue on its way to its destination. To obtain an optimal trajectory for the handover procedure, the potential motion of the mobile robot is essential. The future trajectory of the mobile robot may be predicted based on its recent motion. For instance, neural networks can be used to forecast the observed robot's near future trajectory~\cite{LuoEberhard2022,GiuliariHasanCristaniGalasso2020} or the Bayesian and discrete mechanics and optimal control framework to estimate a long term trajectory~\cite{LuoEberhard2021}. However, the details of the trajectory prediction for the observed mobile robot are not subject of this work, and its future trajectory is presumed to be estimated well enough for planning the trajectory of the aerial manipulator. Hence, in this work, we assume that the aerial manipulator has full prior knowledge of the mobile robot's future trajectory. Given the future or predicted motion of the mobile robot, the proposed method in this study is capable to determine the appropriate handover opportunities, which include the state of the aerial manipulator and the handover time steps. Additionally, the entire travel time for the handover procedure is also to be determined, and it should be as short as possible to minimize the operation time without neglecting the overall system dynamics and its limits.

To address trajectory planning, we present an optimization framework for planning the trajectory of a quadrotor and its onboard manipulator. It takes into account the dynamics of the whole system in order to ensure that the estimated trajectory adheres to the dynamic restrictions. By solving the optimization problem, a time-optimal trajectory is represented with a sequence of states of the quadrotor and its manipulator, while the optimal handover opportunities are determined as well.

Previous research works have proposed approaches for controlling the grasping process itself as well as for planning time-optimal trajectories. An overview is provided subsequently.

\textit{Object Grasping with Onboard Manipulators}: Recent work has examined aerial operation with onboard manipulators in great detail. In our prior work~\cite{LuoEschmannEberhard2022}, a fixed manipulator is attached under the frame of a quadrotor, and an electromagnet has been implemented to grasp a static object on the ground. Although the coupled structure is robust, it lacks the ability to provide any additional motion to handle the grasping object. As a result, the quadrotor cannot engage in aggressive behavior in order to seize its object. In~\cite{ThomasPolinSreenathKumar2013}, a 1-DoF manipulator was used to mimic an avian's grasping trajectory and catch a static cylindrical object at rapid speeds. Both aforementioned studies use a manually planned trajectory represented by polynomials~\cite{MellingerKumar2011}, disregarding the system's dynamics and the effect of the associated mechanics. On the other hand, several research works have also employed complicated manipulators to perform operation tasks. In~\cite{ZhangHeDaiGuYangHanLiuQi2018}, a 7-DoF manipulator is applied to grasp a moving object that provides a larger manipulating space compared to the aforementioned structure setup. The UAV is controlled by a PID-based controller, and the trajectory planning is not discussed in detail.

\textit{Time-Optimal Trajectory Planning}:
The majority of work employs a polynomial-based trajectory to guide UAVs in a mission, as it is straightforward to implement and is continuously differentiable, while the travel time is manually specified based on prior knowledge or experiment. However, due to the restricted flight time, planning a time-optimal trajectory is becoming a popular option for UAVs since it can save operation time and increase the efficiency. In~\cite{VerscheureDemeulenaereSweversDeSchutterDiehl2009,VanLoockPipeleersSwevers2013}, a time-optimal trajectory is generated based on geometric constraints but the quadrotor's dynamics is neglected, meaning that the resulting trajectory might not be realizable by the quadrotor. In~\cite{GaoWuPanZhouShen2018} kinodynamic~\cite{DonaldXavierCannyReif1993} constraints are employed to bound the predicted trajectory. In more recent work~\cite{FoehnRomeroScaramuzza2021}, the authors suggest an optimization strategy, which introduces complementarity constraints to distinguish whether the intended trajectory has traveled through the desired path waypoints in 3D space and the quadrotor's dynamics is also considered in the optimization problem. Furthermore, in~\cite{RomeroSunFoehnScaramuzza2021} a near-time-optimal flight is realized, and the optimal control input is generated to direct the quadrotor passing through path waypoints at high speed.

In contrast, in this work, we propose an optimization-based trajectory planning algorithm for a quadrotor with a 1-DoF manipulator that enables it to approach and hand over a moving target in a time-optimal manner. Instead of manually designing a contact trajectory and arranging the handover states with the moving target, the proposed optimization framework automatically determines contact opportunities that satisfy both the studied quadrotor dynamics with the onboard manipulator and the collaboration with the moving ground mobile robot. The proposed work integrates the discrete mechanics and optimal control (DMOC) framework~\cite{OberBloebaumJungeMarsden2011} and complementarity constraints to create a new framework dubbed Discrete Mechanics and Complementarity Constraints (DMCC). The dynamics of the examined aerial manipulator is not significantly simplified in the proposed optimization framework, which prevents generating an inadmissible trajectory. Furthermore, as inspired in~\cite{FoehnRomeroScaramuzza2021}, the complementarity constraints aid in establishing the optimized handover opportunities throughout the procedure under a variety of desired conditions.

To our best knowledge, this is the first implementation that can handle the handover trajectory planning with a dynamic target for aerial manipulators while considering the full system dynamics and automatically finding the optimal handover opportunities. Moreover, the proposed framework also can be successfully utilized in a classical drone racing trajectory planning, and, as shown below, our framework outperforms the state-of-the-art framework, requiring less computational time. Additionally, the proposed trajectory planning framework is validated not only in simulations but also in real-world experiments with our customized aerial manipulators.

In Section 2, we illustrate some preliminaries about the modeling of the studied aerial manipulator. Then, in Section 3, the proposed discrete mechanics and complementarity constraints trajectory planning framework is introduced, which comprises the discrete mechanics and optimal control framework associated with the aerial manipulator and the complementarity constraints during the handover procedure. Subsequently, several numerical simulations are conducted to verify the performance of our proposed trajectory planning framework in Section 4. In Section 5, the proposed approach is further verified with our aerial manipulator in real-world experiments. Finally, the conclusions and an outlook to future work are given in Section 6.

\section{Preliminaries}
\label{sec:preliminaries}
\subsection{Quadrotor Modeling}
\label{sec:quadrotor_modeling}
At first, the quadrotor itself, without any additional manipulator, is inspected. The system has six degrees of freedom while the number of control inputs is only four, which means that we have an under-actuated system. The state of the quadrotor is defined as
\begin{equation}
  \label{eq:quadrotor_state}
  \bm{x}_\text{g} = \left[\bm{p}^\top \ \bm{\xi}^\top \ \bm{v}^\top \ \bm{\omega}_\text{B}^\top \right]^\top \in \mathbb{R}^{12},
\end{equation}
where the position of the quadrotor is represented by $\bm{p}:=\left[x \ y \ z\right]^\top$ and the velocity is denoted as $\bm{v}=\dot{\bm{p}}:=\left[v_x \ v_y \ v_z \right]^\top$. Moreover, the Euler angles of the quadrotor are denoted as $\bm{\xi}:=\left[\phi \ \theta \ \psi\right]^\top$, and the angular velocity is expressed as $\bm{\omega}_\text{B}$ in the body frame. The input of such a system is represented by the four generated forces from spinning propellers that are denoted as $\bm{u}_\text{g} = \left[f_{\text{motor,1}} \ f_{\text{motor,2}} \ f_{\text{motor,3}} \ f_{\text{motor,4}}\right]^\top$, resulting in a collective mass-normalized thrust $f_\text{thrust} = \frac{1}{m_\text{quadrotor}} \sum_{n = 1}^{4}f_{\text{motor,i}}$ along the $z$-axis of the body-fixed frame of the quadrotor and a torque $\bm{\tau}$ that is calculated as
\begin{equation}
  \label{eq:torque_motors}
  \bm{\tau}=\begin{bmatrix}
    \frac{\sqrt{2}}{4}\ell_\text{frame} (f_\text{motor,2}+f_\text{motor,3}-f_\text{motor,1}-f_\text{motor,4}) \vspace{1mm} \\ \frac{\sqrt{2}}{4}\ell_\text{frame} (f_\text{motor,2}+f_\text{motor,4}-f_\text{motor,1}-f_\text{motor,3}) \vspace{1mm}\\ c_\tau (f_\text{motor,3}+f_\text{motor,4}-f_\text{motor,1}-f_\text{motor,2})
  \end{bmatrix},
\end{equation}
where the diagonal length of the quadrotor frame is denoted as $\ell_\text{frame}$ and the torque coefficient is denoted as $c_\tau$. Given the notation above, the general nonlinear quadrotor dynamics is formulated as
\begin{equation}
  \label{eq:quad_system_dyanmics}
  \dot{\bm{x}}_\text{g}  = \bm{f}(\bm{x}_\text{g}, \bm{u}_\text{g})= \begin{bmatrix}
    \bm{v} \\ \bm{T}(\bm{\xi})\bm{\omega}_\text{B} \\ \bm{R}_\text{B}(\bm{\xi}) \left[0 \ 0 \ f_\text{thrust}\right]^\top - \left[0 \ 0 \ g\right]^\top \\ \bm{J}_\text{quadrotor}^{-1}(-\bm{\omega}_\text{B}\times\bm{J}_\text{quadrotor}\bm{\omega}_\text{B} + \bm{\tau})
  \end{bmatrix},
\end{equation}
where the gravitational acceleration is denoted as $g=9.8066~$m/s$^2$, the matrix $\bm{J}_\text{quadrotor}$ denotes the quadrotor's moment of inertia, and the rotation matrix $\bm{R}_\text{B}(\bm{\xi})$ denotes the transformation from the body frame to the inertial frame of reference. Furthermore, the mapping matrix between the angular velocities between the body frame and the inertial frame is defined as
\begin{equation*}
  \bm{T}(\bm{\xi}) = \left[\begin{array}{ccc}
      1 & \sin(\phi)\tan(\theta) & \cos(\phi)\tan(\theta) \\
      0 & \cos(\phi)             & -\sin(\phi)            \\ 0& \frac{\sin(\phi)}{\cos(\theta)} & \frac{\cos(\phi)}{\cos(\theta)}
    \end{array}\right].
\end{equation*}

\subsection{Aerial Manipulator Modeling}
The dynamics definition in Eq.~\eqref{eq:quad_system_dyanmics} reveals the substantial nonlinear characteristics of the quadrotor. When one implements an additional manipulator, the system becomes even more sophisticated. Previously published work has usually taken a decentralized approach, in which the flying platform and its attached mechanical components or payloads are considered independently~\cite{RuggieroLippielloOllero2018}. Under these circumstances, the quadrotor and its additional components including the payload dynamics are not considered in a coupled manner; instead, the influence of the additional components is usually regarded as a disturbance~\cite{RuggieroCacaceSadeghianLippiello2014}. On the other hand, one can estimate the dynamics of the whole system using either the Euler-Lagrangian or the Newton-Euler approach. In some early works using the Newton-Euler formulations~\cite{ZhangHeDaiGuYangHanLiuQi2018,MasoneBuelthoffStegagno2016,SonSeoKimJinKim2018}, several simplifications are applied to decouple the system dynamics; on the contrary, the Euler-Lagrangian approaches described in~\cite{RuggieroLippielloOllero2018,SamadikhoshkhoGhorbaniJanabiSharifi2021} handle the UAV and its payload as a single system without manually and explicitly considering the interaction between each body.

\begin{figure}[htpb]
  \centering
  \includegraphics[scale=1]{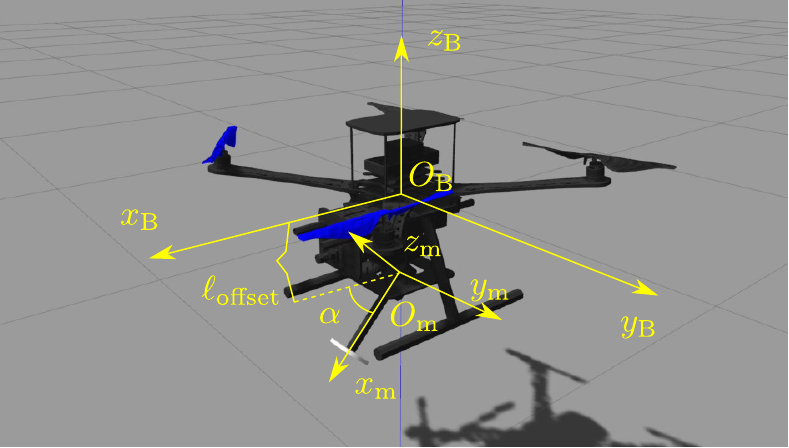}
  \caption{Coordinates of the aerial manipulator. The coordinates of the quadrotor and the manipulator are denoted as $(\cdot)_\text{B}$ and $(\cdot)_\text{m}$, respectively.}
  \label{fig:coordinate_body}
\end{figure}

In this work, considering to propose a time-optimal trajectory for pushing the aerial manipulator to its limit, the dynamics of the aerial manipulator is derived following a discrete variational Euler-Lagrangian approach without significant simplifications. The studied aerial manipulator is composed of a quadrotor and a 1-DoF manipulator that is driven by a servo motor, and the angle of the manipulator with respect to the $x$-axis of the quadrotor's body frame is represented with $\alpha$, as shown in Fig.~\ref{fig:coordinate_body}. The end-effector is a permanent magnet, whose structural properties can be disregarded here. Besides, the offset between the mounting point of the manipulator and the mass center of the quadrotor is denoted as $\bm{\ell}_\text{offset, B}$, which as illustrated in Fig.~\ref{fig:coordinate_body} is represented in the quadrotor frame as $\left[0 \ 0 \ -\ell_\text{offset}\right]^\top$. The kinematic and dynamic modeling is described in the following.

\textit{Kinematic modeling}:
Given the state of the quadrotor defined in Eq.~\ref{eq:quadrotor_state}, the manipulator's mass center position $\bm{p}_\text{m}$ is denoted as
\begin{equation}
  \label{eq:position_manipulator}
  \begin{aligned}
    \bm{p}_\text{arm} = & \ \bm{p}+ \bm{R}_\text{B}(\bm{\xi})  \left(  \bm{\ell}_\text{offset,B} + \ ^{\text{B}}\bm{R}_\text{m}(\alpha) \begin{bmatrix}\frac{\ell_\text{arm}}{2}\\0\\0 \end{bmatrix} \right), \\
                        & \text{with} \hspace{1mm} ^{\text{B}}\bm{R}_\text{m}(\alpha) = \left[\begin{array}{ccc}
                                                                                                  \cos(\alpha) & 0 & \sin(\alpha) \\
                                                                                                  0            & 1 & 0            \\ -\sin(\alpha)& 0 & \cos{\alpha}
                                                                                                \end{array}\right],
  \end{aligned}
\end{equation}
where the length of the onboard manipulator is denoted as $\ell_\text{arm}$, and we assume that the center of mass is identical of the geometric center of the manipulator. Similarly, the position of the end-effector of the manipulator $\bm{p}_\text{end-effector}$ can be calculated by
\begin{equation}
  \label{eq:position_manipulator_endeffector}
  \bm{p}_\text{end-effector} = \bm{p}+ \bm{R}_\text{B}(\bm{\xi})  \left(  \bm{\ell}_\text{offset,B} +\ ^{\text{B}}\bm{R}_\text{m}(\alpha) \begin{bmatrix}\ell_\text{arm}\\0\\0 \end{bmatrix} \right).
\end{equation}
Furthermore, the translational velocity of the manipulator center $\bm{v}_\text{arm}$ can be calculated as
\begin{equation}
  \label{eq:velocity_manipulator}
  \begin{aligned}
    \bm{v}_\text{arm} = & \ \bm{v}  - \dfrac{\dot{\alpha}\ell_\text{arm}}{2}\bm{R}_\text{B}(\bm{\xi}) \begin{bmatrix} \sin(\alpha)\\0\\ \cos(\alpha) \end{bmatrix} \\ & + \bm{R}_\text{B}(\bm{\xi}) \bm{\omega}_\text{B} \times  \left(  \bm{\ell}_\text{offset,B}  + \dfrac{\ell_\text{arm}}{2}\begin{bmatrix} \cos(\alpha)\\0\\ -\sin(\alpha) \end{bmatrix}\right),
  \end{aligned}
\end{equation}
where the angular velocity of the servo motor is denoted as $\dot{\alpha}$. Additionally, the angular velocity of the manipulator represented in the inertial frame can be calculated as
\begin{equation}
  \label{eq:angular_velocity_manipulator_b}
  \dot{\bm{\xi}}_\text{arm} = \bm{R}_\text{B}(\bm{\xi})\left( \ ^{\text{B}}\bm{R}_\text{m}(\alpha) \begin{bmatrix} 0\\\dot{\alpha}\\0 \end{bmatrix} + \bm{\omega}_\text{B}\right).
\end{equation}

\textit{Dynamic modeling}:
\label{sec:dynamic_modeling}
In Lagrangian mechanics, a mechanical system is described in terms of generalized coordinates, which in this work are specified as $\bm{q}:=\left[x \ y \ z \ \phi \ \theta \ \psi \ \alpha \right]^\top = \left[\bm{p}^\top \ \bm{\xi}^\top \ \alpha \right]^\top$ for the studied aerial manipulator. Then, the Lagrange function for the aerial manipulator is defined as
\begin{equation}
  \label{eq:lagrangian}
  L(\bm{q}, \dot{\bm{q}})= K(\bm{q}, \dot{\bm{q}}) - V(\bm{q}),
\end{equation}
where $\dot{\bm{q}}$ denotes the time derivative accordingly, and the kinetic and potential energies of the aerial manipulator are denoted as $K$ and $V$, respectively.

The kinetic energy can be further specified as
\begin{equation}
  \label{eq:kinetic_energy}
  \begin{aligned}
    K(\bm{q}, \dot{\bm{q}}) = & \ \dfrac{m_\text{quadrotor}}{2}\bm{v}^\top\bm{v} + \dfrac{1}{2}\bm{\omega}_\text{B}^\top \bm{J}_\text{quadrotor}\bm{\omega}_\text{B} \\ & + \dfrac{m_\text{arm}}{2}\bm{v}_\text{arm}^\top\bm{v}_\text{arm}+ \dfrac{1}{2}\dot{\bm{\xi}}_\text{arm,m}^\top \bm{J}_\text{arm}\dot{\bm{\xi}}_\text{arm,m},
  \end{aligned}
\end{equation}
where the masses of the quadrotor and its onboard manipulator are denoted as $m_\text{quadrotor}$ and $m_\text{arm}$, respectively, and $\bm{J}_\text{arm}$ represents the inertia of the manipulator respecting to each body-fixed frame shown as $O_\text{m}$ in Fig.~\ref{fig:coordinate_body}. The angular velocity of the manipulator is transformed into the coordinates of its body-fixed rotation axis
\begin{equation}
  \dot{\bm{\xi}}_\text{arm,m} = \ ^\text{B}\bm{R}_\text{m}^\top(\alpha)\bm{R}_\text{B}^\top(\bm{\xi})\dot{\bm{\xi}}_\text{arm}.
\end{equation}
Then, the potential energy of the aerial manipulator is given by
\begin{equation}
  \label{eq:potential_energy}
  V(\bm{q}) = m_\text{quadrotor}gz + m_\text{arm}g\bm{p}_\text{m}^\top\left[0 \ 0 \ 1\right]^\top.
\end{equation}

Furthermore, an arbitrary motion of such a system can be represented by a curve along its configuration manifold $\mathbb{Q}$~\cite{McLachlanPerlmutter2006}, which begins from an initial state $(\bm{q}(t_0), \dot{\bm{q}}(t_0))$ to an end state $(\bm{q}(t_N),\dot{\bm{q}}(t_N))$ under the influence of control forces throughout the time interval $\left[0, \ t_N\right]$. In continuous mechanics, the integration of the Lagrange function $L$ in this particular trajectory $\bm{q}(t)$ is called the Hamilton action $\mathfrak{S}$
\begin{equation}
  \label{eq:hamilton_action}
  \mathfrak{S} =\int_0^{t_N} L(\bm{q}(t), \dot{\bm{q}}(t))\text{d}t.
\end{equation}
According to Hamilton's principle, among all feasible trajectories connecting the specified initial and final conditions at $t=0$ and $t=t_N$, the true trajectories are those that make $\mathfrak{S}$ stationary. Thus, the motion of the mechanical system between these two time stamps yields
\begin{equation}
  \label{eq:least_action}
  \delta \mathfrak{S} = 0,
\end{equation}
where the variation of the Hamilton action is denoted as $\delta \mathfrak{S}$. Thereby, Hamilton's principle is also known as the least action principle, since the action is a local minimum for a true trajectory.

Finally, based on Eq.~\eqref{eq:least_action}, one obtains
\begin{equation}
  \label{eq:euler_lagrangian_equation}
  \dfrac{\text{d}}{\text{d}t}\dfrac{\partial L}{\partial \dot{\bm{q}}} - \dfrac{\partial L}{\partial \bm{q}} = \bm{0},
\end{equation}
which are also called the Euler-Lagrange equations.

When considering non-conservative forces $f_\text{e}$, the least-action principle from Eq.~\eqref{eq:least_action} can be described as
\begin{equation}
  \label{eq:lagrange_dalembert}
  \delta \int_0^{t_N} L(\bm{q}(t), \dot{\bm{q}}(t))\text{d}t + \int_0^{t_N}f_\text{e}(\bm{q}(t), \dot{\bm{q}}(t))\delta \bm{q} \text{d}t = 0,
\end{equation}
which is known as the Lagrange-d'Alembert principle.

\section{Time-Optimal Trajectory Planning}
\subsection{Trajectory Planing Problem}
The method proposed in this work shall automatically establish a time-optimal trajectory for guiding an aerial manipulator towards and grasping a dynamic target. There are several requirements that the planned trajectory should fulfill. The end position of the manipulator $\bm{p}_{\text{end-effector}}$ should approach the handover target $\bm{p}_{\text{target}}$ at least once during the procedure, giving the end-effector a sufficient time window to grasp the target object and complete the handover task. Remark that this time window should be determined given the potential motion of the dynamic target without any manual arrangement. The total travel time of the entire procedure $t_N$ should be minimized to improve the cooperation efficiency. Furthermore, the relative velocity and the moving direction between the manipulator and the mobile robot must be limited during the handover in accordance with particular experimental criteria or hardware restrictions. In any case, the planned trajectory should not exceed the dynamic limit of the aerial manipulator.

\subsection{DMOC Framework}
\label{sec:dmoc_framework}
In general, the procedure for incorporating the system dynamics into the trajectory generation problem optimization-based begins with deriving ordinary differential equations from the Newton-Euler equations or the Euler-Lagrangian defined in Eq.~\eqref{eq:euler_lagrangian_equation}, and then discretizing the dynamics using methods such as the 4-th order explicit Runge-Kutta method (RK4)~\cite{FoehnRomeroScaramuzza2021,SunBergAlterovitz2016}. On the other hand, the system dynamics can be derived using the discrete variational principle~\cite{WendlandtMarsden1997}, and this formulation also can be utilized as a numerical integrator in the optimization problem. In this approach, the discretization of Lagrange-d'Alembert principle occurs first and then the equations derived from the variational principle are employed to determine the constraints describing the system dynamics.

Based on the second approach, the discrete mechanics and optimal control (DMOC) framework, first suggested in~\cite{JungeMarsdenOberBloebaum2005}, can be utilized to plan the optimal trajectories of non-holonomic mobile robots~\cite{KobilarovSukhatme2007}, manipulators~\cite{OberBloebaumJungeMarsden2011}, satellites~\cite{JungeMarsdenOberBloebaum2005} and quadrotors~\cite{ZuZhangShan2017} that are subject to the system dynamics. The DMOC can even be utilized to solve an optimal control problem, with a performance comparable to that of a standard nonlinear model predictive control framework~\cite{XuTimmermannTraechtler2017,IsmailLiu2018}. Thus, in this work, we utilize the DMOC framework to formulate the aerial manipulator's dynamics constraints as well.

In discrete variational mechanics, the trajectory in the time interval $\left[0, \ t_N\right]$ is first discretized. The discrete configuration is described with a vector $\bm{q}_k\in\mathbb{Q}$ using generalized coordinates described the system's configuration at the time steps $t_k=k\Delta t, k\in\left[0, \ldots, N\right]$. Moreover, the planned total travel time can be obtained by $t_N:=N\Delta t$.

Given the aforementioned discretization, based on the discrete variational principle~\cite{Marsden1999}, the continuous configuration space $\mathbb{Q}$ and its tangent configuration space $\mathbb{TQ}$ are approximated by the discrete configuration space $\mathbb{Q}\times\mathbb{Q}$, and the discrete form of the Lagrange function $L$ is denoted as $L_\text{d}$,
\begin{equation}
  \label{eq:discrete_Lagrangian_approx}
  L_\text{d}(\bm{q}_k, \bm{q}_{k+1}) \approx \int_{t_k}^{t_{k+1}}L(\bm{q}(t), \dot{\bm{q}}(t))\text{d}t,
\end{equation}
which approximates the action over a time interval between $t_k$ and $t_{k+1}$. Then, based on Eqs.~\eqref{eq:hamilton_action} and~\eqref{eq:discrete_Lagrangian_approx} the discrete Hamilton action can be calculated as
\begin{equation}
  \mathfrak{S}_\text{d} = \sum_{k=0}^{N} L_\text{d}(\bm{q}_k, \bm{q}_{k+1}),
\end{equation}
and its variation should be zero according to Hamilton's principle. Furthermore, the integration of the virtual work is approximated by
\begin{equation}
  \sum_{k=0}^{N} \bm{f}_k^{-}\delta \bm{q}_k + \bm{f}_k^{+}\delta \bm{q}_{k+1} \approx \int_0^{t_N} f_\text{e}(\bm{q}(t), \dot{\bm{q}}(t))\delta \bm{q} \text{d}t,
\end{equation}
where the external forces are represented at each discrete time step by the left/right forces, denoted by $\bm{f}^{-}_k$ and $\bm{f}^+_k$, respectively~\cite{MarsdenWest2001}. Particularly, as introduced in~\cite{JungeMarsdenOberBloebaum2005} both discrete forces can be treated equally around each time step as
\begin{equation}
  \label{eq:discrete_forces}
  \bm{f}_k^- = \bm{f}_k^+     = \frac{\Delta t}4\left(\bm{f}_{k+1}+\bm{f}_{k}\right).
\end{equation}

Thereby, the discrete forced Euler-Lagrange equations can be derived for each consecutive time interval, yielding
\begin{equation}
  \label{eq:discrete_lagrange}
  \dfrac{\partial L_\text{d}(\bm{q}_{k-1},\bm{q}_k)}{\partial \bm{q}_k}  + \dfrac{\partial L_\text{d}(\bm{q}_k,\bm{q}_{k+1})}{\partial\bm{q}_k} + \bm{f}_{k-1}^+ +  \bm{f}_{k}^-  = \bm{0},
\end{equation}
where the index $k$ is in the interval $\{1, \ldots, N-1\}$. The boundary conditions for the ending positions can be deduced using the discrete Legendre transforms~\cite{OberBloebaumJungeMarsden2011} defined as
\begin{equation}
  \label{eq:discrete_impulse}
  \begin{aligned}
    \mathfrak{p}_k     & = \dfrac{\partial L_d(\bm{q}_{k-1},\bm{q}_k)}{\partial\bm{q}_k} + \bm{f}_{k-1}^{+},       \\
    \mathfrak{p}_{k-1} & = -\dfrac{\partial L_d(\bm{q}_{k-1},\bm{q}_k)}{\partial\bm{q}_{k-1}} - \bm{f}_{k-1}^{-},
  \end{aligned}
\end{equation}
where $\mathfrak{p}_k$ indicates the discrete generalized impulse at the time step $t_k$. Then, Eq.~\eqref{eq:discrete_impulse} is used to approximate the known generalized impulse $\mathfrak{p}(t_k)$ at the ending time step at $t_0$ and $t_N$ that can be calculated through
\begin{equation}
  \label{eq:impluse}
  \mathfrak{p}(t_k) := \dfrac{\partial L(\bm{q},\dot{\bm{q}})}{\partial\dot{\bm{q}}}\bigg|_{{\scriptstyle \bm{q} = \bm{q}(t_k)}, \ {\scriptstyle \dot{\bm{q}} = \dot{\bm{q}}(t_k)}}.
\end{equation}
Based on Eqs.~\eqref{eq:discrete_impulse} and~\eqref{eq:impluse}, the boundary conditions are defined as
\begin{equation}
  \label{eq:dmoc_boundary_conditions}
  \begin{aligned}
     & \dfrac{\partial L(\bm{q},\dot{\bm{q}})}{\partial\dot{\bm{q}}}\bigg|_{\begin{aligned} & {\scriptstyle \bm{q} = \bm{q}(t_0)}\\[-1ex]   & {\scriptstyle \dot{\bm{q}} = \dot{\bm{q}}(t_0)}\end{aligned}} + \dfrac{\partial L_d(\bm{q}_0,\bm{q}_1)}{\partial\bm{q}_0} + \bm{f}_0^{-} =\bm{0},      \vspace{5mm} \\
     & -\dfrac{\partial L(\bm{q},\dot{\bm{q}})}{\partial\dot{\bm{q}}}\bigg|_{\begin{aligned} &{\scriptstyle \bm{q} = \bm{q}(t_N)} \\[-1ex] & {\scriptstyle \dot{\bm{q}} = \dot{\bm{q}}(t_N)}\end{aligned}} + \dfrac{\partial L_d(\bm{q}_{N-1},\bm{q}_N)}{\partial\bm{q}_N} + \bm{f}_{N-1}^{+} =\bm{0},
  \end{aligned}
\end{equation}
given the initial and final conditions of the system $\bm{q}(t_0)$, $\dot{\bm{q}}(t_0)$, $\bm{q}(t_N)$, and $\dot{\bm{q}}(t_N)$, respectively.

Besides, there are various implementations of $L_\text{d}(\bm{q}_k, \bm{q}_{k+1})$ that can be chosen. In this work, the Verlet method is selected, which has been applied first in the research of molecular dynamics in~\cite{Verlet1967}. It is straightforward to be implemented in the optimization problems, and provides several attractive numerical properties. Based on the Verlet method, the discrete Lagrange function is defined as
\begin{equation}
  \label{eq:verlet_lagrangeian}
  L_\text{d}(\bm{q}_k, \bm{q}_{k+1}) = \frac{1}{2}\Delta tL\left( \bm{q}_k,\bm{\dot{q}}_{k,k+1} \right) + \frac{1}{2}\Delta tL\left( \bm{q}_{k+1}, \bm{\dot{q}}_{k, k+1}\right),
\end{equation}
where the average velocity between the two time steps is assumed to be constant and calculated by $\bm{\dot{q}}_{k,k+1} = (\bm{q}_{k+1}-\bm{q}_k)/\Delta t$.

Finally, based on the equations above, the optimization problem based on the DMOC framework is denoted as follow
\begin{equation}
  \label{eq:dmoc_framework}
  \begin{aligned}
    \min_{t_N^*,\bm{u}^*,\bm{q}^*} & \hspace{3mm} J_\text{d}(t_N,\bm{u}_k,\bm{q}_k)                                                                                                  \\
    \text{s.t.}                    & \hspace{3mm} \Delta t = t_N/N,                                                                                                                  \\
                                   & \hspace{3mm}\bm{q}_0 = \bm{q}(t_0) \hspace{2mm}\text{and}  \hspace{2mm}\dot{\bm{q}}_0 = \dot{\bm{q}}(t_0),                                      \\
                                   & \hspace{3mm}\bm{q}_N = \bm{q}(t_N) \hspace{2mm}\text{and}  \hspace{2mm}\dot{\bm{q}}_N = \dot{\bm{q}}(t_N), \hspace{2mm}                         \\ &\hspace{3mm}\bm{q}_{\text{min}} \leq \bm{q}_{k} \leq \bm{q}_{\text{max}}, \hspace{1mm} \forall  k\in \{0, \ldots, N\}, \\
                                   & \hspace{3mm}\dot{\bm{q}}_{\text{min}} \leq \bm{\dot{q}}_{k,k+1} \leq \dot{\bm{q}}_{\text{max}},  \hspace{1mm} \forall  k\in \{0, \ldots, N-1\}, \\
                                   & \hspace{3mm}\bm{u}_{\text{min}} \leq \bm{u}_k \leq \bm{u}_{\text{max}},  \hspace{1mm} \forall  k\in \{0, \ldots, N-1\},                         \\
                                   & \hspace{3mm}\text{discrete Euler-Lagrange equations based on}                                                                                   \\ &\hspace{3mm}\text{Eqs.}~\eqref{eq:discrete_forces}- \eqref{eq:verlet_lagrangeian},
  \end{aligned}
\end{equation}
where $J_\text{d}$ denotes the discrete objective function of the optimization problem, and the control input of the system at each time step is represented with $\bm{u}_k$. In this work, the control input is composed of four generated forces from motors and the torque of the servo motor that drives the motion of the manipulator. Hence, the external forces are identical to the control input.

\subsection{Complementarity Constraints}
\label{sec:complementarity_constraints}
Apart from adhering to system dynamics and satisfying the ending conditions, the intended trajectory should drive the aerial manipulator to approach its dynamic target and accomplish the handover procedure. Thus, within the planned trajectory, the end-effector of the aerial manipulator should have a period of time in which it is close enough to the ground mobile robot to hand over the target. However, the timing of this handover is hard to be arranged manually for a dynamic target, and is strongly coupled with the system dynamics. Moreover, the desired behavior of the aerial manipulator during the handover and the non-contact phases may differ significantly. For instance, the relative velocity between the end-effector of the manipulator and the moving target should be decreased during the handover to guarantee a steady contact, while both velocities are completely decoupled during the non-contact phases. Thus, by solving the optimization problem, one should be able to distinguish whether the aerial manipulator is in the handover phase at an arbitrary time step $t_k$.

To indicate the handover phase as well as the handover opportunities, we introduce a set of progress index variables $\epsilon_k, \text{where} \ k\in \{0, \ldots, N-1\}$, as inspired by the work from~\cite{FoehnRomeroScaramuzza2021}. In our design, each progress index variable $\epsilon_k$ highlights the handover state at each discrete time step $t_k$, and it should be non-zero if and only if our desired contact conditions are fulfilled, which are to be detailed later. Hence, if $\epsilon_k$ is zero, there is no planned contact between the manipulator's end-effector and the object at time step $t_k$.

To implement this design and distinguish the handover timing mathematically, one may formulate it as
\begin{equation}
  \label{eq:cc_definition}
  \epsilon_k f_\text{c}(z) = 0, \hspace{3mm} \text{with} \hspace{3mm} f_\text{c}(z)\geq0,
\end{equation}
or marked equivalently with the complementarity operator $\bot$ as
\begin{equation}
  \label{eq:cc_definition_simple}
  0\leq  \epsilon_k\bot f_\text{c}(z) \geq 0,
\end{equation}
where $f_\text{c}(z)$ is the specified non-negative condition functions of some state $z$. Equations~\eqref{eq:cc_definition} and~\eqref{eq:cc_definition_simple} are also known as complementarity constraints, where the relationship between the progress index $\epsilon_k$ and the condition functions $f_\text{c}(z)$ is specified so that at least one of them must remain zero to hold the equality constraint.

According to this design, the first and most straightforward contact condition is the judge of the Euclidean distance between the end-effector of the manipulator and the target, which is defined as
\begin{equation}
  \label{eq:contact_condition}
  0\leq \epsilon_k \bot\parallel \bm{p}_{\text{end-effector}, k} - \bm{p}_{\text{target},k}\parallel \geq0,
\end{equation}
where $\parallel\cdot\parallel$ denotes the Euclidean norm of the given vector. When the end-effector is in the situation of touching and grasping the target, the progress index can be non-zero; on the other phases, the progress index variable should remain zero. Ideally each progress index variable $\epsilon_k$ should be binary so that it can unambiguously identify the handover time steps. For example, when the progress index variable equals to one, it indicates that the distance condition is fulfilled and the manipulator is grasping the target; on the contrary, the progress index variable remains zero so that the distance condition can be disregarded in this constraint~\eqref{eq:contact_condition}. However, this would lead to a mixed integer problem, which is notoriously difficult to solve. Thus, in this work the range of each progress index variable $\epsilon_k$ is relaxed and set to be in the interval $\left[0.0, 1.0\right]$.

Nevertheless, solely based on condition~\eqref{eq:contact_condition}, the handover opportunities may not be indicated, since the progress index can remain at zero no matter which value of the Euclidean norm has been calculated during the whole optimization procedure. To avoid this situation, an additional set of the progress variables $\bm{\kappa}$ is introduced, and at each time step the progress index variable $\epsilon_k$ must satisfy
\begin{equation}
  \label{eq:progress_condition}
  \begin{aligned}
    \epsilon_k  & = \kappa_{k} - \kappa_{k+1}, \hspace{2mm} k \in \{0, \ldots, N-1\},                          \\
    \text{with} & \hspace{2mm} \kappa_0 = \kappa_\text{init} \hspace{2mm}\text{and}\hspace{2mm} \kappa_N = 0,
  \end{aligned}
\end{equation}
where the positive value of $\kappa_\text{init}$ is specified by the user. Thus, due to $\kappa_\text{init}$ being positive and the final condition of $\kappa_N=0$, the progress index variable $\epsilon_k$ must be non-zero at a number of time steps. Since $\epsilon_k \in \left[0, 1\right]$, the number of time steps with $\epsilon_k \neq 0$ is at least $\kappa_\text{init}$. Therefore, as $\kappa_\text{init}$ increases, the minimum planned contact duration increases.

Remark that unlike the setup based on the similar mechanism for trajectory planning to traverse multiple predefined path waypoints in~\cite{FoehnRomeroScaramuzza2021,LuoEberhard2021}, where the initial progress variable $\kappa_\text{init}$ is set to one since each intended path waypoint should be passed through only once, in this work this initial value can be set up with a reasonable positive number. The rationale for that is the handover is supposed not to be accomplished solely within one single time step, since a rash grasping makes a solid handover procedure impossible. On the contrary, the handover duration is able to be adjusted using this initial value $\kappa_\text{init}$, so that the planned trajectory has a sufficient amount of handover time to accomplish the task.

\begin{figure}[htpb]
  \centering
  \includegraphics[scale=1]{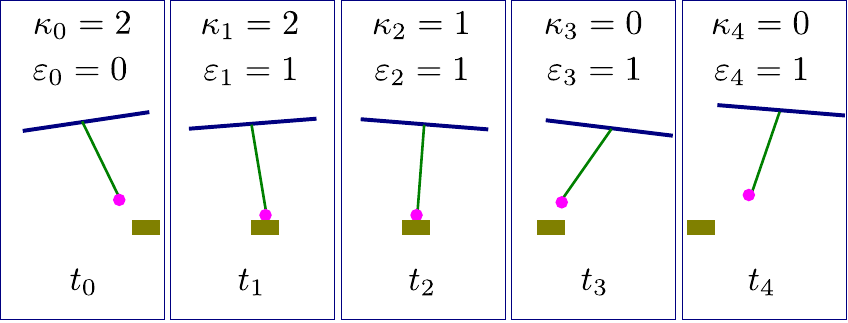}
  \caption{Illustration of the handover progress with $\kappa_\text{init}=2$. The frame of the quadrotor is shown with a blue line, and the green line indicates the manipulator, while the pink circle marking the position of the end-effector. Moreover, the position of the moving grasping target is illustrated with olive squares.}
  \label{fig:cc_illustration}
\end{figure}

In Fig.~\ref{fig:cc_illustration}, a handover demonstration is shown, where the change of both the progress index variables and the progress variables of the produced handover procedure is illustrated with the corresponding time step. In this scenario, the initial value of $\kappa_\text{init}$ is set to two, indicating that there are two progress index variables $\epsilon_k$ should be one; meanwhile in the planned trajectory the manipulator should maintain contact for grasping at least two time steps. Based on the designed progress conditions in Eq.~\eqref{eq:progress_condition} and the contact condition defined in~\eqref{eq:contact_condition}, at time steps $t_1$ and $t_2$, the progress index variables become 1 since the end-effector has gotten close enough to the moving grasping target. Note that the results presented here are intended to demonstrate the relationship between the progress variables and the progress index variables; results for planned trajectories will be discussed in further detail in the next section.

Based on the same mechanism introduced above, one can engage more conditions that must be satisfied during the contact phase. In this work, we introduce two additional constraints based on the complementarity constraints. On the one hand, the relative velocity between the end-effector and the handover target should be minimized, so that the impact between the end-effector and the handover target can be reduced. The formulation of this constraint is defined as
\begin{equation}
  \label{eq:speed_cc}
  0\leq \epsilon_k \bot\parallel \bm{v}_{\text{end-effector}, k} - \bm{v}_{\text{target},k}\parallel \geq0.
\end{equation}
On the other hand, the heading of the quadrotor is set identical to the moving direction of the moving grasping object during the handover procedure, so that the handover procedure will not be disturbed by the landing gears. The heading constraint during the handover procedure is specified as
\begin{equation}
  \label{eq:heading_cc}
  0\leq \epsilon_k \bot\left\lvert \bm{v}_{\text{target},k}^{xy} \times \bm{x}_{\text{B},k}^{xy}\right\rvert \geq0, \hspace{2mm} k \in \{0, \ldots, N-1\},
\end{equation}
where $\left\lvert\cdot\right\rvert$ denotes the absolute value. Furthermore, $\left(\cdot\right)^{xy}$ denotes the first two components of the given vector, and the vector $\bm{x}_{\text{B},k}$ denotes the $x$-axis of the quadrotor that is represented in the inertial frame of reference at the time step $t_k$.

\subsection{DMCC Formulation}
In this work, the optimization problem is formulated based on the introduced DMOC framework in~\eqref{eq:dmoc_framework}, together with the complementarity constraints discussed in Section~\ref{sec:complementarity_constraints}. However, directly incorporating the complementarity constraints into the optimization problem as an MPCC, such as in~\cite{FoehnFalangaKuppuswamyTedrakeScaramuzza2017}, may make it more difficult to solve since the classical constraint qualifications are failed to hold due to the complementarity constraints~\cite{KanzowSchwartz2013}. Typically, the MPCC problem needs to be reformulated, and the complementarity constraints may be relaxed with some numerical strategies. As shown in~\cite{FoehnRomeroScaramuzza2021,LuoEberhard2021}, one can introduce an additional relaxation variable set into the complementarity constraints to mitigate the effect arising from the complementarity constraints. Thereby, the complementarity constraints defined in~\eqref{eq:contact_condition} is reformulated as
\begin{equation}
  \label{eq:re_contact_condition}
  0\leq \epsilon_k \bot(\parallel \bm{p}_{\text{end-effector}, k} - \bm{p}_{\text{target},k}\parallel-\nu_k) \geq0,
\end{equation}
where the relaxation variable $\nu_k$ is defined to be positive within a small value. Except numerical requirements, in real-world applications, a slight offset between the position of the end-effector and the handover target is likely acceptable, given that we employ a permanent magnet to grasp the target and the magnetic force can exert its influence at a specific distance.

For the remaining complementarity constraints defined in~\eqref{eq:speed_cc} and~\eqref{eq:heading_cc}, another implementation strategy is adopted. As discussed in~\cite{BaumruckerRenfroBiegler2008}, the equilibrium conditions can be relaxed and replaced with inequalities, such as
\begin{equation}
  \epsilon_k f_\text{c}(z) =0\Longrightarrow  \epsilon_kf_\text{c}(z) \leq c_\text{limit},
\end{equation}
where $c_\text{limit}$ denotes a small constant parameter, e.g., $10^{-2}$.

Moreover, to plan a smooth trajectory and avoid unnecessary aggressive actions throughout the procedure, the cost function penalizes the difference between the optimized control input and the reference control input at each time step. Based on the introduction above, the proposed optimization framework DMCC is formulated as
\begin{equation}
  \label{eq:final_optimization}
  \begin{aligned}
    \min_{t_N^*,\bm{u}^*,\bm{q}^*} \hspace{3mm} & J_\text{d} =  t_N +c_u\Delta t\sum_{k=0}^N \parallel\bm{u}_k - \bm{u}_\text{ref}\parallel                                      \\
    \text{s.t.}   \hspace{3mm}                  & \epsilon_k (\parallel \bm{p}_{\text{end-effector}, k} - \bm{p}_{\text{target},k}\parallel-\nu_k) = 0,                          \\
                                                & \epsilon_k \parallel \bm{v}_{\text{end-effector}, k} - \bm{v}_{\text{target},k}\parallel \leq c_\text{limit, velocity},        \\
                                                & \epsilon_k \left\lvert \bm{v}_{\text{target},k}^{xy} \times \bm{x}_{\text{B},k}^{xy}\right\rvert \leq c_\text{limit, heading}, \\
                                                & \nu_k \in \left[0, \nu_\text{max}\right],                                                                                      \\
                                                & \text{progress iteration defined in Eq.~\eqref{eq:progress_condition}},                                                        \\
                                                & \text{constraints from the DMOC framework~\eqref{eq:dmoc_framework}},
  \end{aligned}
\end{equation}
where $\bm{u}_\text{ref}$ denotes the reference control input, which in this work is defined as the control input for the aerial manipulator hovering in the air without rotating its manipulator as $\left[f_\text{hover}\, f_\text{hover}\, f_\text{hover}\, f_\text{hover}\, 0\right]^\top$, where the hover force generated from each motor amounts to $f_\text{hover}=0.25(m_\text{quadrotor}+m_\text{arm})g$. The penalization parameter $c_u$ is selected as 3$\cdot10^{-3}$.

\section{Numerical Simulations}
In this section, the performance of the proposed trajectory planing framework DMCC is demonstrated in several scenarios to verify its performance. The proposed framework is implemented in Python with the CasADi library~\cite{AnderssonGillisHornRawlingsDiehl2019} in conjunction with the solver IPOPT~\cite{WaechterBiegler2005}.

\subsection{Trajectory Planning for Aerial Races}
One of the challenging applications for the aerial robots is aerial racing, where the UAV travels with its maximal velocity to pass through several static checkpoints in the shortest time. The authors from~\cite{FoehnRomeroScaramuzza2021} show the state of the art performance of trajectory planning while considering the quadrotor's dynamics through several static gateways in a 3D space. Their results can even defeat the top human UAV racers. In~\cite{FoehnRomeroScaramuzza2021}, the authors used standard Newton-Euler equations to formulate the dynamics constraints and discretized them with the RK4 method. In contrast, we implement dynamic constraints using the proposed DMCC framework based on~\eqref{eq:final_optimization} to plan the time-optimal trajectory for the quadrotor in this work.

\begin{figure}[htpb]
  \centering
  \includegraphics[scale=0.9]{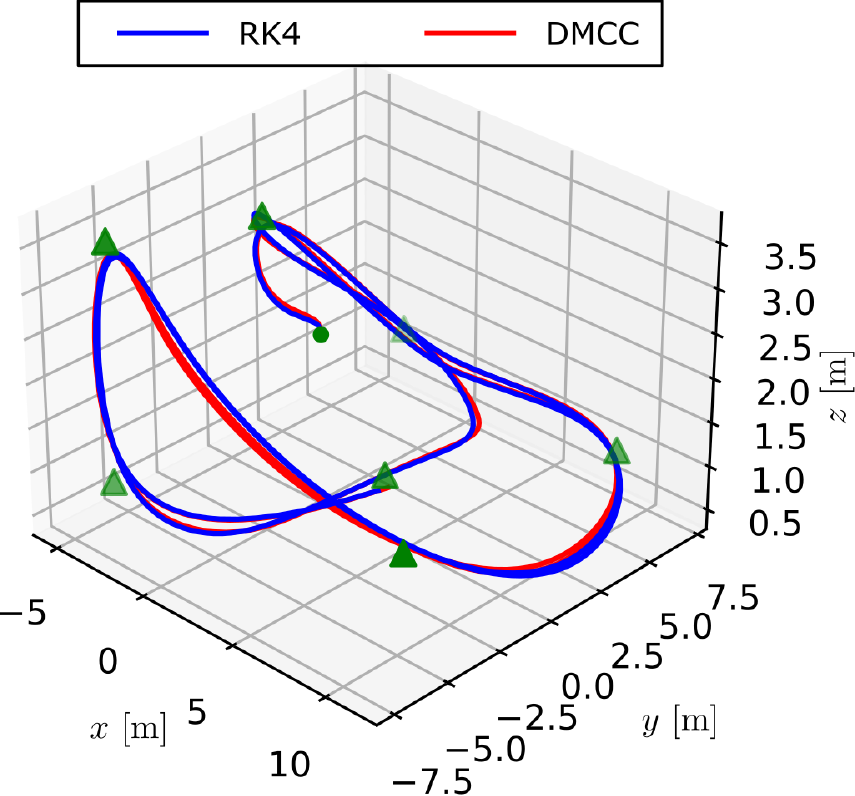}
  \caption{Generated racing trajectory}
  \label{fig:generated_racing_trajectory}
\end{figure}

In the proposed racing scenario from~\cite{FoehnRomeroScaramuzza2021}, several path waypoints are predefined, as indicated by the green triangles in Fig.~\ref{fig:generated_racing_trajectory}. These waypoints should be traversed by the intended path in the correct order. Despite the different optimization problem formulations with different numerical integrators, the remaining conditions and parameters, such as the initial state of the quadrotor and the physical parameters of the quadrotor, are kept the same.

\begin{figure}[htpb]
  \centering
  \includegraphics[scale=0.8]{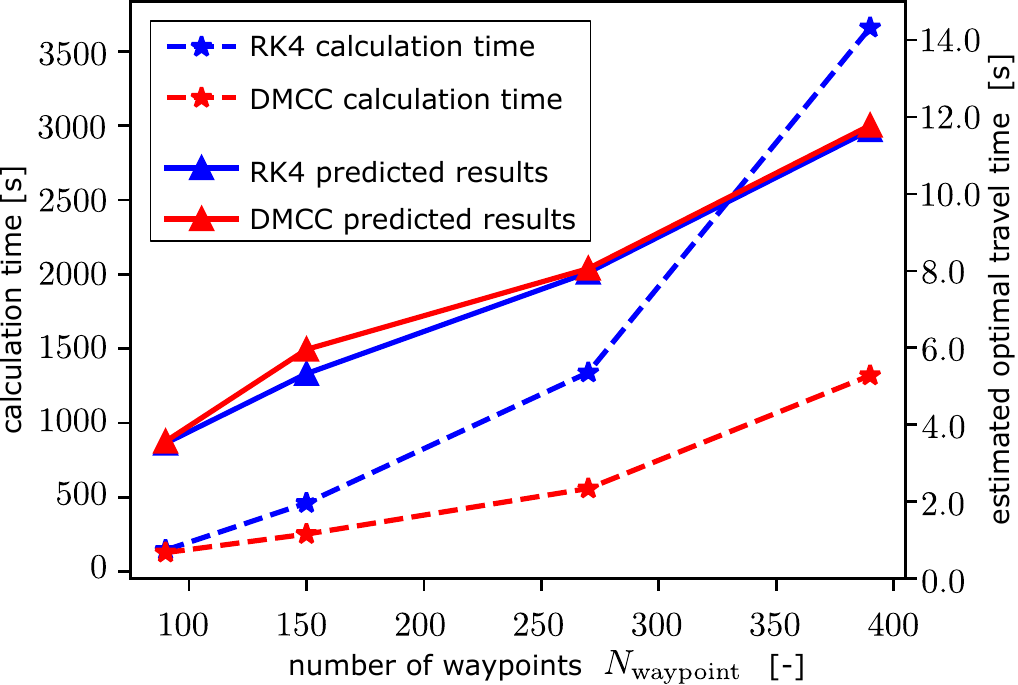}
  \caption{Performance comparison between the RK4-based approach~\cite{FoehnRomeroScaramuzza2021} and the DMCC framework. The results are measured on a machine equipped with an Intel Core i9-8950HK CPU and 32 GB RAM.}
  \label{fig:comparsion_rk4_dmcc}
\end{figure}

Although the resulting trajectories in Fig.~\ref{fig:generated_racing_trajectory} are comparable, the proposed DMCC framework exhibits a superior numerical performance, see Fig.~\ref{fig:comparsion_rk4_dmcc}. With more path waypoints, it is straightforward to deduce that the corresponding optimal travel time will grow as the path lengthens. Both frameworks provide very close predicted optimal travel time. However, the consumption of the computational time shows significant difference. Obviously, more path waypoints necessitate more variables in the optimization problem, which makes it more difficult for the solver to find an appropriate solution. As the number of path waypoints rises, the proposed framework demands less computational time to obtain a comparable performance compared to the work from~\cite{FoehnRomeroScaramuzza2021}.

\subsection{Handover Trajectory Planning}
\begin{figure*}[!t]
  \centering
  \subfloat[handover at a fixed position]{\includegraphics[scale=1]{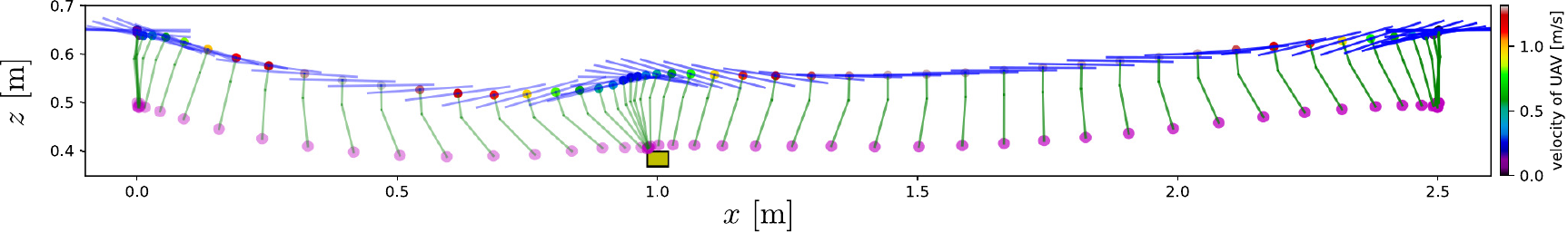}%
    \label{fig:ns_fixed_xz}}
  \hfil
  \subfloat[handover with a linear moving target]{\includegraphics[scale=1]{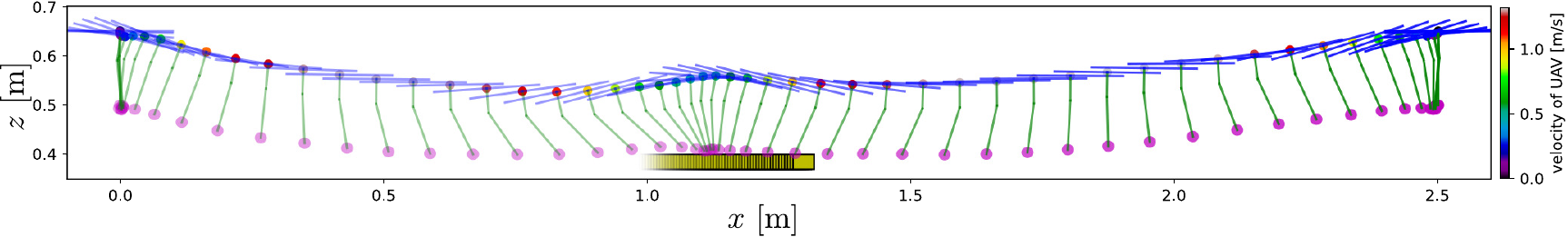}%
    \label{fig:ns_linear_xz}}\hfil
  \subfloat[handover with a target that is moving with a circular trajectory (in the $x$-$z$-plane) ]{\includegraphics[scale=1]{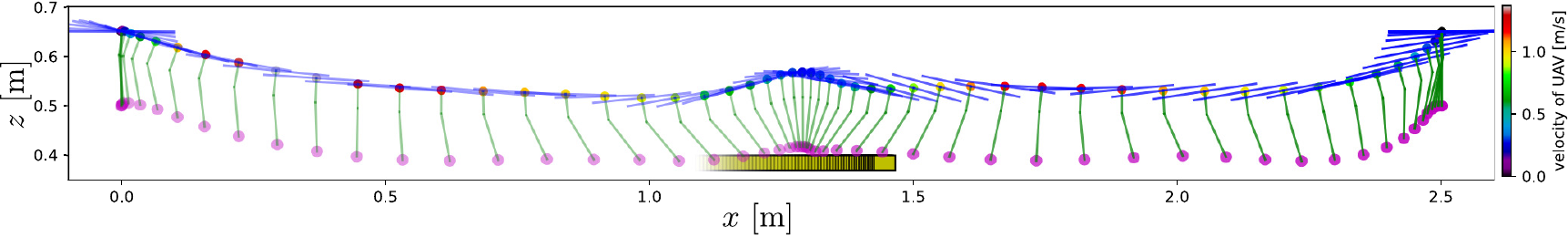}%
    \label{fig:ns_circle_xz}}
  \hfil
  \subfloat[handover with a target that is moving with a circular trajectory (in the $x$-$y$-plane)]{\includegraphics[scale=1]{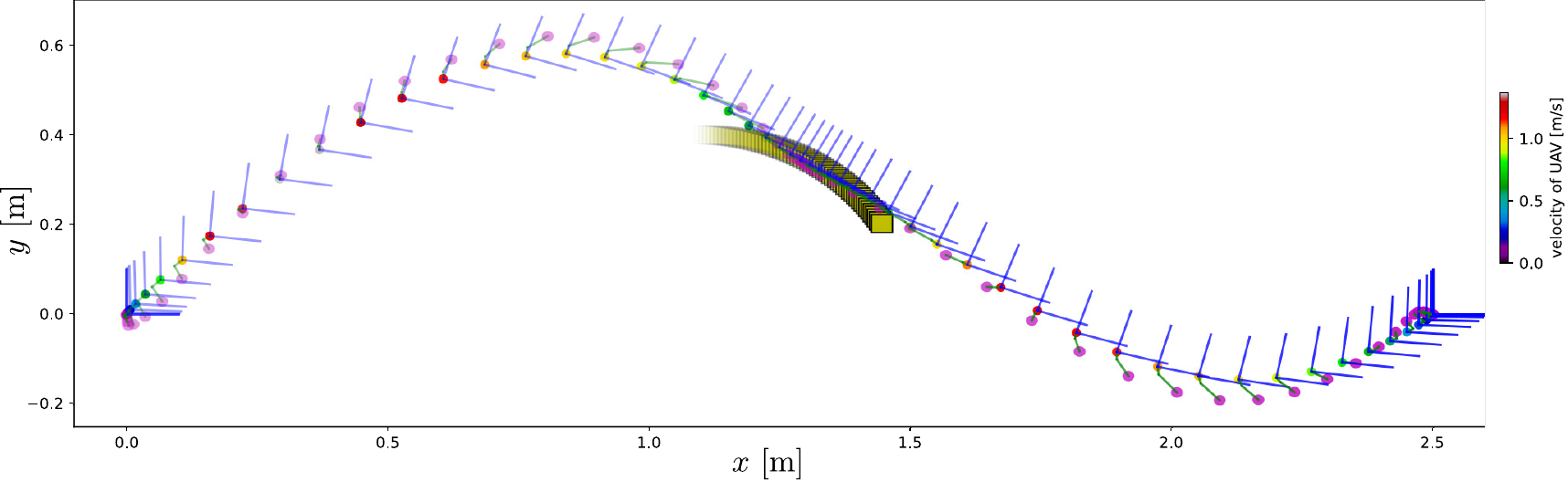}%
    \label{fig:ns_circle_xy}}
  \caption{Three scenarios of the handover cooperations}
  \label{fig:ns_results}
\end{figure*}

In this section, several dynamic handover scenarios are illustrated to validate the proposed DMCC framework, see Fig.~\ref{fig:ns_results}. In particular, the grasping target is not always in a fixed location. Compared to the aerial races in the previous section, the dynamic system is more complex due to an additional manipulator that is mounted under the frame of the quadrotor. In each scenario, the aerial manipulator begins from the initial position $\begin{bmatrix}0.0 \ 0.0 \ 0.65\end{bmatrix}^\top$~m, and ends at the final position $\left[2.5 \ 0.0 \ 0.65\right]^\top$~m. We assume that the end-effector has a magnetic component and that the grasping object is magnetically attractable. The maximum acceptable offset between the end-effector and the grasping object $\nu_\text{max}$ is set to be 0.02~m, assuming that the magnet is strong enough. The other simulation parameters are presented in Tab.~\ref{tab:numerical_simulation_parameters}.

\begin{table}[!t]
  \caption{Numerical Simulation Parameters}

  \centering
  \begin{tabular}{|m{3.6cm}|c|}
    \hline
    \centering
    parameters                                                                                                  & value                                                 \\
    \hline
    \centering
    \vspace{1mm}UAV mass $m_\text{quadrotor}\vspace{1mm}$                                                       & 1.659~[kg]                                            \\
    \hline
    \centering
    \vspace{1mm}manipulator mass  $m_\text{arm}$                          \vspace{1mm}                          & 0.36~[kg]                                             \\
    \hline
    \centering
    \vspace{1mm} quadrotor inertia \vspace{1mm}    $\bm{J}_\text{quadrotor}$                                    & diag$\left(0.0348, 0.0459, 0.0977\right)$~[kg\,m$^2$] \\
    \hline
    \centering
    \vspace{1mm}manipulator inertia \vspace{1mm}     $\bm{J}_\text{arm}$                                        & diag$\left(0.0, 0.0019, 0.0\right)~$[kg\,m$^2$]       \\
    \hline\centering
    \vspace{1mm}manipulator  offset $\bm{\ell}_\text{offset,B}\vspace{1mm}$                                     & $\left[0 \ 0 \ -0.05\right]^\top$~[m]                 \\
    \hline\centering
    \vspace{1mm}initial $\kappa$ $\kappa_\text{init}$                          \vspace{1mm}                     & 2                                                     \\
    \hline\centering
    \vspace{1mm}max. allowed offset between end-effector and the grasping target $\nu_\text{max}$  \vspace{1mm} & 0.02~[m]                                              \\
    \hline\centering
    \vspace{1mm}max. allowed velocity difference by handover $c_\text{limit, velocity}$      \vspace{1mm}       & 0.01~[m/s]                                            \\
    \hline\centering
    \vspace{1mm}max. allowed heading difference by handover $c_\text{limit, heading}$      \vspace{1mm}         & 0.1~[rad]                                             \\
    \hline\centering
    \vspace{1mm}max./min. velocity               \vspace{1mm}                                                   & $\left[\pm1.3 \ \pm1.3 \ \pm1.15\right]^\top$~[m/s]   \\
    \hline\centering
    \vspace{1mm}max./min. angular velocity           \vspace{1mm}                                               & $\left[\pm8.0 \ \pm8.0 \ \pm2.0 \right]^\top$~[rad/s] \\
    \hline\centering
    \vspace{1mm}max./min. manipulator rotation velocity                           \vspace{1mm}                  & $\pm \pi/2$~[rad/s]                                   \\
    \hline
  \end{tabular}
  \label{tab:numerical_simulation_parameters}
\end{table}

\begin{figure}[htpb]
  \centering
  \includegraphics[scale=0.78]{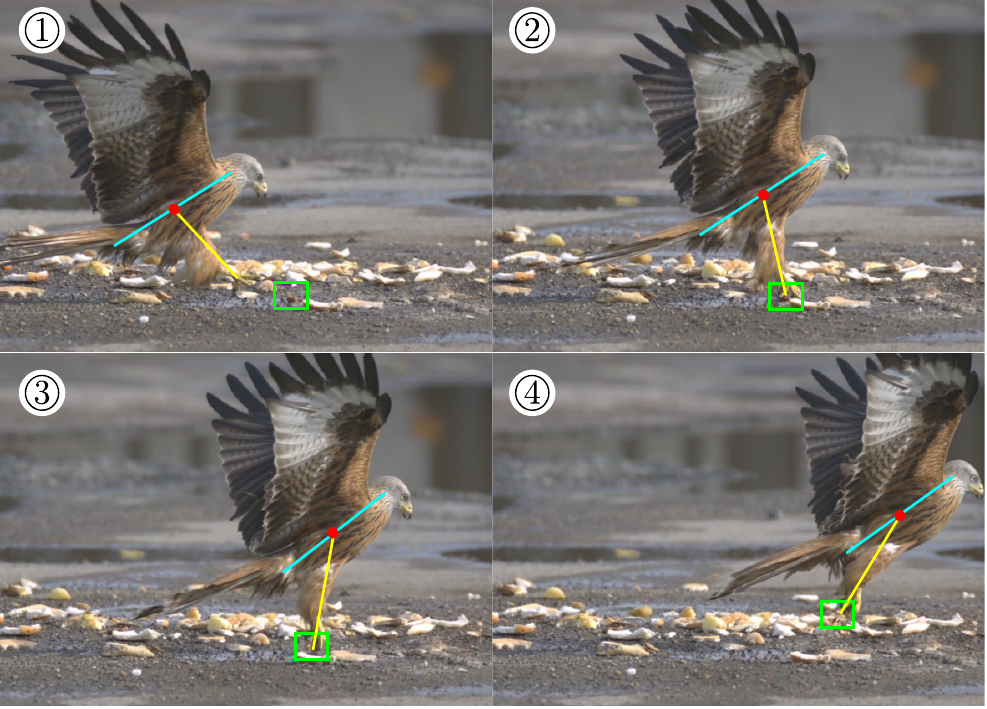}
  \caption{Still pictures from video footage of a red kite\protect\footnotemark}
  \label{fig:hawk_images}
\end{figure}

\subsubsection{Static Grasping Object}
During the handover operation in the first scenario, the handover target remains at the fixed position $\begin{bmatrix}1.0 \ 0.0 \ 0.4\end{bmatrix}^\top$~m. In a previous work~\cite{ThomasPolinSreenathKumar2013}, a similar scene is set up, but they first studied the motion of a red kit from a series of video footages, such as Fig.~\ref{fig:hawk_images}, and then imitate the red kite's grasping strategy by generating a polynomial trajectory. On the contrary, in this study, the proposed framework solely involves the system dynamics and the position of the target to automatically generate a dynamics-involved trajectory illustrated in Fig.~\ref{fig:ns_fixed_xz}, where the frame projection of the quadrotor on the $x$-$z$-plane is illustrated with blue lines and the manipulator is simplified with green lines, while its end-effector is indicated by pink circles.

Compared to the the recorded video footage, it is noteworthy to remark that the conducted trajectory in this work shares some similarities compared to the real behavior of the red kit. Based on the planned trajectory in Fig.~\ref{fig:ns_fixed_xz}, the manipulator performs as the claw of the eagle that swings forward to come near to the object prior to grasping it (see Fig.~\ref{fig:hawk_images} \raisebox{.5pt}{\textcircled{\raisebox{-.9pt} {1}}}), then swings back to increase the contact period and reduce the relative velocity with the target when grasping (see Fig.~\ref{fig:hawk_images} \raisebox{.5pt}{\textcircled{\raisebox{-.9pt} {4}}}). Furthermore, given the positive initial progress variable $\kappa_\text{init}$, the end-effector has a period of contact time that it maintains contact with the object in Fig.~\ref{fig:ns_fixed_xz}, exhibiting similar behavior compared to the video footages \raisebox{.5pt}{\textcircled{\raisebox{-.9pt} {2}}} and \raisebox{.5pt}{\textcircled{\raisebox{-.9pt} {3}}} in Fig.~\ref{fig:hawk_images}.

\begin{figure}[htpb]
  \centering
  \subfloat[end-effector trajectory in the $z$ direction]{\includegraphics[scale=1]{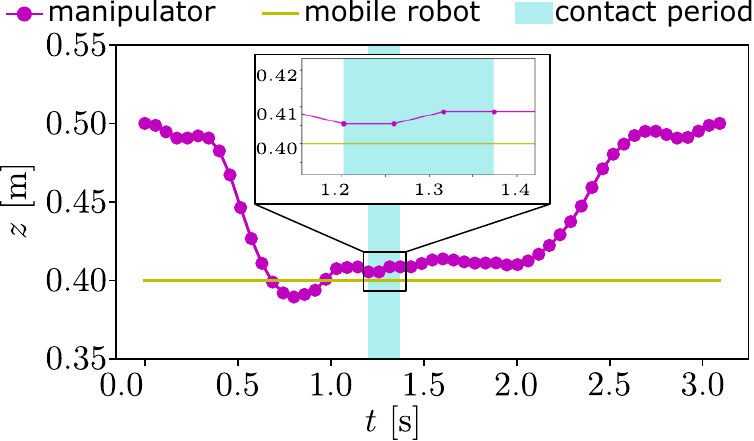}%
    \label{fig:ns_fixedpoint_manipulator_target_z}}
  \hfil
  \subfloat[end-effector velocity in the $x$ direction]{\includegraphics[scale=1]{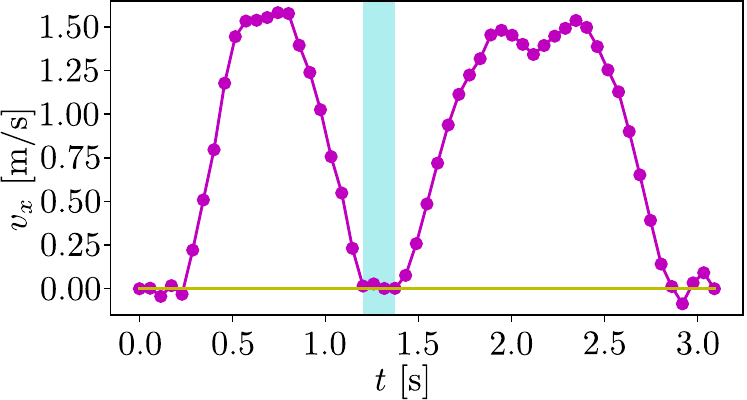}%
    \label{fig:ns_fixedpoint_manipulator_vx}}
  \caption{End-effector trajectory and velocity during the handover procedure with a static object.}
  \label{fig:fixed_point_manipulator_target_z}
\end{figure}

In Fig.~\ref{fig:ns_fixedpoint_manipulator_target_z}, the trace of the end-effector is compared to the position of the grasping object, and the contact period, denoted by cyan hues, is determined by the estimated results from $\epsilon$. Note that although the initial value of $\kappa_\text{init}$ is set to 2 as shown in Tab.~\ref{tab:numerical_simulation_parameters}, the contact period has more than two time steps where the end-effector remains in touch with the object for grasping, since the process index variable is not established as a binary variable. During the contact period, the end-effector stays within the permissible contact range, which is defined by $\nu_\text{max}$ in Tab.~\ref{tab:numerical_simulation_parameters}. Furthermore, due to the design of the optimization condition in~\eqref{eq:speed_cc}, the end-effector's velocity during the grasping is nearly zero, minimizing the relative motion between the end-effector and the grasping object, as shown in Fig.~\ref{fig:ns_fixedpoint_manipulator_vx}.

\subsubsection{Linearly Moving Object}
\begin{figure}[htpb]
  \centering
  \subfloat[end-effector trajectory in the $z$ direction]{\includegraphics[scale=1]{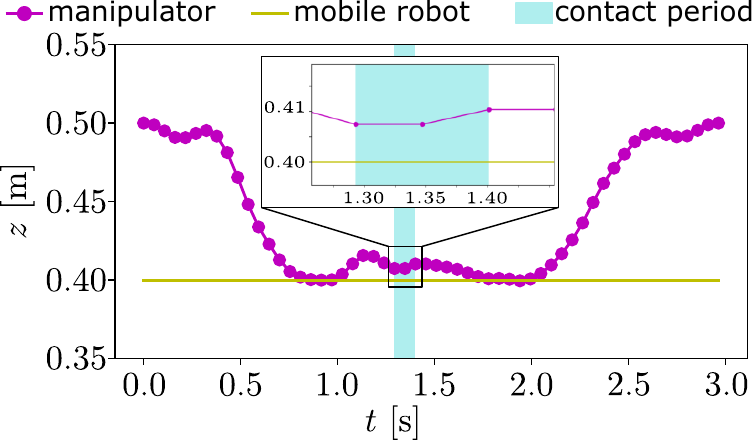}%
    \label{fig:ns_linear_manipulator_z}}
  \hfil
  \subfloat[end-effector velocity in the $x$ direction]{\includegraphics[scale=1]{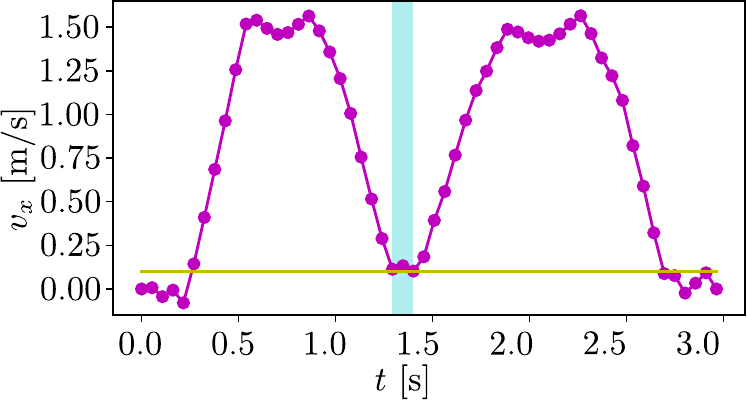}%
    \label{fig:ns_linear_manipulator_vx}}
  \caption{End-effector trajectory and velocity during the handover procedure with a linearly moving object.}
  \label{fig:ns_linear}
\end{figure}

In the second scenario, the mobile robot with the grasping object moves with the velocity 0.1~m/s in the $x$ direction rather than staying at a fixed position. The estimated trajectory is illustrated in Fig.~\ref{fig:ns_linear_xz}, which has a similar performance compared to the planned trajectory for a static object. As illustrated in Fig.~\ref{fig:ns_linear_manipulator_z}, the end-effector has descended its altitude to ensure the grasping within the acceptable limits, while its velocity is made consistent with the moving target around 0.1~m/s in Fig.~\ref{fig:ns_linear_manipulator_vx}. Due to the fact that the velocity of the aerial manipulator does not have to reduce to zero to satisfy the grasping velocity limits, the travel time of the optimal trajectory in this scene can be slightly less than the one in the previous situation.

\footnotetext{The Slow Mo Guys 2011 Red Kites in Slow Motion http://youtu.be/AYOx-iCMZhk}
\subsubsection{Circular Moving Object}
A more challenging scenario is provided where the motion direction of the grasping target changes during the handover procedure. The mobile robot is moving in a circular trajectory with a diameter of 0.4~m and a constant altitude of $z(t)=0.4$~m, which begins from $\left[1.1 \ 0.4 \ 0.4\right]^\top$ and is defined as
\begin{equation}
  \label{eq:ns_circle_trajectory}
  x(t)=0.4\sin(0.3t)+1.1, \hspace{2mm} y(t)=0.4\cos(0.3t).
\end{equation}
As defined complementarity constraints based on inequalities~\eqref{eq:heading_cc}, the planned motion of the aerial manipulator should change its heading to correspond to the motion direction of the mobile robot while remaining the speed and contact condition in the desired constraints defined in~\eqref{eq:contact_condition} and~\eqref{eq:speed_cc}. In Fig.~\ref{fig:ns_circle_xz}, the action of the manipulator is similar to the last previous scenarios, which swings forward first and then backward to ensure a sufficient grasping timing. Besides, the aerial manipulator is turning and adjusts its heading to be similar to the motion direction of the mobile robot until the grasping procedure is accomplished as illustrated in Fig.~\ref{fig:ns_circle_xy}.

\begin{figure}[htpb]
  \centering
  \subfloat[end-effector trajectory in the $z$ direction]{\includegraphics[scale=1]{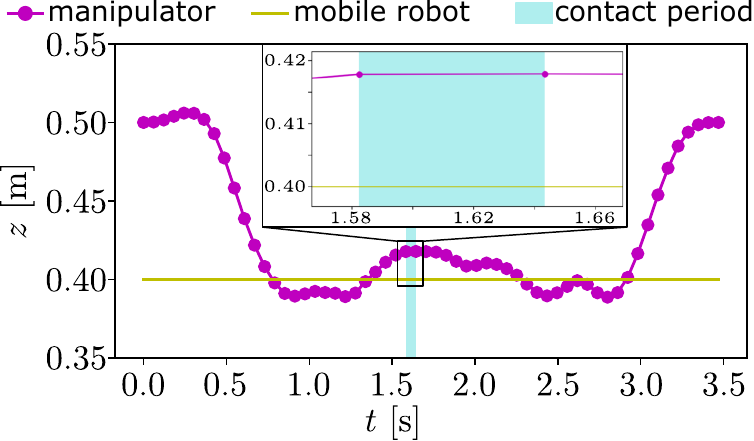}%
    \label{fig:ns_circle_manipulator_xz}}
  \hfil
  \subfloat[end-effector velocity in the $x$ direction]{\includegraphics[scale=1]{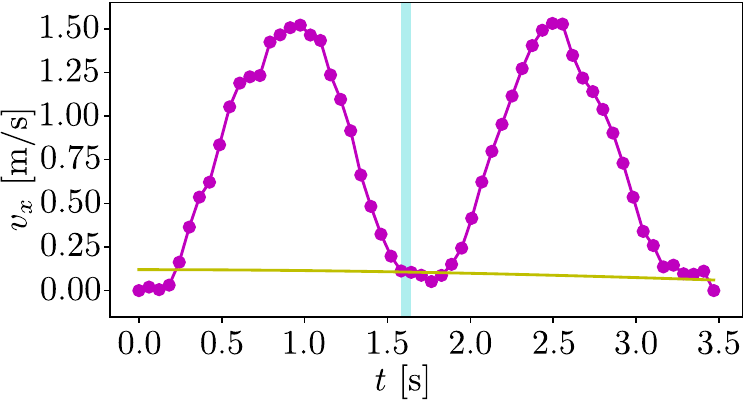}%
    \label{fig:ns_circle_manipulator_vx}} \hfil
  \subfloat[end-effector velocity in the $y$ direction]{\includegraphics[scale=1]{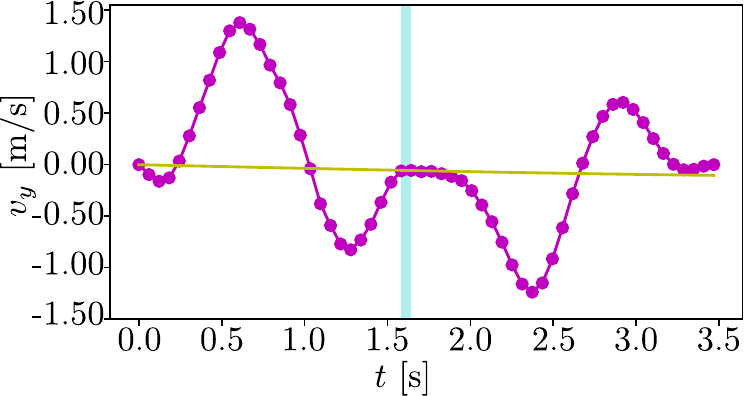}%
    \label{fig:ns_circle_manipulator_vy}}\hfil
  \subfloat[quadrotor heading during the procedure]{\includegraphics[scale=1]{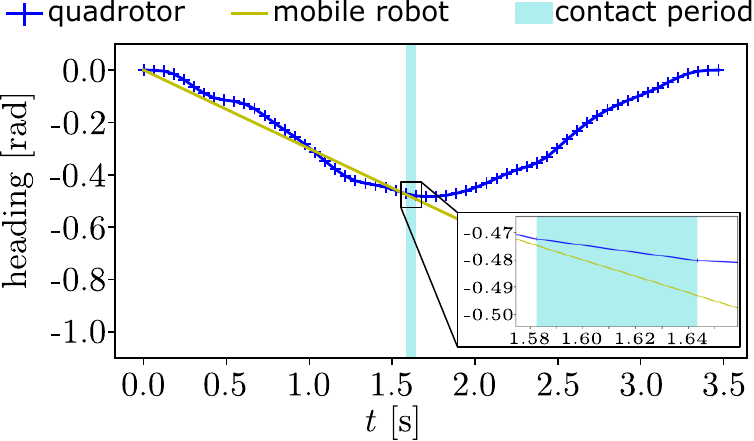}%
    \label{fig:ns_circle_heading}}
  \caption{Trajectories during the handover procedure with an object moving along a circular trajectory.}
  \label{fig:ns_circle}
\end{figure}

The details of the proposed trajectory are further illustrated in Fig.~\ref{fig:ns_circle}. The contact altitude of the manipulator is constrained, so that the manipulator can grasp the target during the handover period. The manipulator velocities in both the $x$ and $y$ directions are illustrated in Figs.~\ref{fig:ns_circle_manipulator_vx} and~\ref{fig:ns_circle_manipulator_vy}. During the contact period, the velocity of the manipulator is reduced to fit the motion of the mobile robot due to the velocity constraints~\eqref{eq:speed_cc}. As expected in Fig.~\ref{fig:ns_circle_heading}, the heading is adjusted to the motion direction of the moving target during the contact period.

As demonstrated by numerical simulations, the proposed approach can determine the optimal trajectory for aerial manipulators. During the contact phase, the position of the end-effector corresponds well with that of the grasping target. The same applies for velocity and heading. Without any manual involvement or assistance, the proposed DMCC can provide time-optimal solutions to the trajectory planning problems for nonlinear systems.

\section{Experiments}
\subsection{Control Framework}
\begin{figure*}[htpb]
  \centering
  \includegraphics[scale=1]{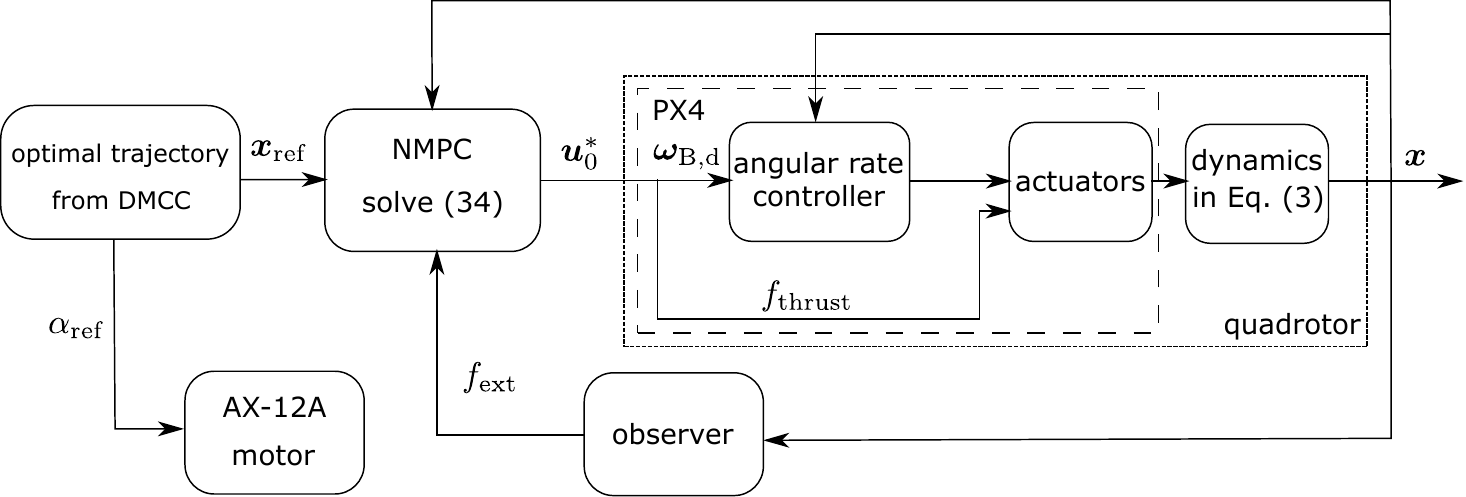}
  \caption{Control framework}
  \label{fig:control_framework}
\end{figure*}

In this section, the estimated trajectory is utilized to direct our self-developed aerial manipulator to grasp an object in a dynamic environment. Besides the proposed time-optimal trajectory generator, an appropriate control framework is required. In this work, we utilize a nonlinear model predictive controller (NMPC) to control the quadrotor to follow the desired optimal trajectory in 3D space. To reduce the computational complexity of the optimization problem to be solved for the NMPC controller, the onboard manipulator is controlled separately, namely using a model-based feed-forward controller. The proposed control framework is illustrated in Fig.~\ref{fig:control_framework}.

Additionally, in real-world experiments, we utilize a Pixhawk flight control unit to regulate the angular velocities of the onboard rotors. Thus, the force $f_{\text{motor},i}$ generated from the $i$-th motor is not controlled directly by our proposed NMPC algorithm. Instead, the Pixhawk flight control unit can track the desired angular rates using feedback from onboard gyroscopes and an indoor localization system. Moreover, compared to the aerial manipulator's moment of inertia, the spinning propellers can generate large torques to rotate the whole aerial manipulator, which results in a fast response to desired angular rate commands. Thereby, as illustrated in Fig.~\ref{fig:control_framework}, the quadrotor's rotational dynamics is neglected in the NMPC control design, i.e.~the NMPC controller assumes that it can directly govern the quadrotor's angular velocities. Therefore, the control input introduced in Section~\ref{sec:dynamic_modeling} is converted from $\bm{u}_\text{g}$ to $\bm{u}=\left[\bm{\omega}_\text{B,d}^\top \ f_\text{thrust}\right]^\top$, where the desired angular rate of the quadrotor is denoted as $\bm{\omega}_\text{B,d}:=\left[\omega_{x} \ \omega_{y} \ \omega_{z}\right]^\top$.

Moreover, to avoid the singularities in Eq.~\eqref{eq:quad_system_dyanmics}, a normalized quaternion $\hat{\bm{q}}:=\left[q_w \ q_x \ q_y \ q_z\right]^\top$ is employed to describe the orientation of the quadrotor rather than the Euler angles. Thus, the state of the quadrotor is redefined as $\bm{x}:=\left[\bm{p}^\top \ \hat{\bm{q}}^\top \ \bm{v}^\top\right]^\top$. Moreover, the time derivative of the quaternion is calculated by
\begin{equation}
  \dot{\hat{\bm{q}}} = \dfrac{1}{2} \bm{Q}(\bm{\omega}_\text{B,d})\hat{\bm{q}},
\end{equation}
where $\bm{Q}(\bm{\omega}_\text{B,d})$ denotes a skew-symmetric matrix from $\bm{\omega}_\text{B,d}$~\cite{FalangaFoehnLuScaramuzza2018}, which is defined as
\begin{equation}
  \bm{Q}(\bm{\omega}_\text{B,d}) = \left[\begin{array}{cccc}
      0        & -\omega_x & -\omega_y & -\omega_z \\
      \omega_x & 0         & \omega_z  & -\omega_y \\
      \omega_y & -\omega_z & 0         & \omega_x  \\
      \omega_z & \omega_y  & -\omega_x & 0
    \end{array}\right].
\end{equation}
Finally, the dynamics of the quadrotor for the NMPC controller is then described as
\begin{equation}
  \label{eq:nominal_dyns}
  \dot{\bm{x}}=f_\text{nominal}(\bm{x}, \bm{u})=\begin{bmatrix}\dot{\bm{p}}\\ \dot{\hat{\bm{q}}}  \\ \dot{\bm{v}}\end{bmatrix} =\begin{bmatrix}
    \bm{v}                                                  \\
    \dfrac{1}{2} \bm{Q}(\bm{\omega}_\text{B,d})\hat{\bm{q}} \\[2ex]
    \bm{R}_\text{B}(\hat{\bm{q}})  \begin{bmatrix}0 \\ 0 \\ f_\text{thrust}\end{bmatrix} - \begin{bmatrix}0 \\ 0 \\ g\end{bmatrix}
  \end{bmatrix},
\end{equation}
where the matrix $\bm{R}_\text{B}(\hat{\bm{q}})$ denotes the rotation transformation given the quaternion~\cite{NekooAcostaOllero2022}.

Furthermore, due to unknown aerodynamics, unbalanced mass distributions, and other disturbances that are challenging to model directly, the actual quadrotor dynamics in real-world experiments is not identical to the nominal model described in Eq.~\eqref{eq:nominal_dyns}. Since model-based controllers derive their control solutions from models, the model discrepancy between the nominal model and the real one may have a negative impact on control performance. Thus, to improve the model-based controller's performance, similar to our previous work~\cite{LuoEschmannEberhard2022}, the system dynamics is augmented with an additional data-driven disturbance observer that is based on a trained Gaussian process regression (GPR) model. The augmented system dynamics is defined as
\begin{equation}
  f_\text{augmented}(\bm{x},\bm{u}) = f_\text{nominal}(\bm{x},\bm{u}) + f_\text{ext}(\bm{x}),
\end{equation}
where the model difference predicted by the GPR model is denoted as $f_\text{ext}(\bm{x})$. The GPR model generates the predicted disturbances to account for model differences while observing the specified vector $\bm{z}$ of the aerial manipulator at each time step. In this work, the observed vector $\bm{z}:=\left[z \ \alpha \ v_{\text{B},x} \ v_{\text{B},y} \ v_{\text{B},z}\right]$ corresponds to the altitude of the aerial manipulator, the angle of the manipulator, and the velocities of the quadrotor that are represented in its body-fixed frame $O_\text{B}$. Based on the observed vector, the model difference given the training data can be estimated by
\begin{equation}
  f_\text{ext}(\bm{x}) = \bm{B}_\text{z}\bm{R}_\text{B}(\hat{\bm{q}})\bm{\mu}(\bm{z})= \bm{B}_\text{z}\bm{R}_\text{B}(\hat{\bm{q}}) \begin{bmatrix} \mu_x(z, \alpha, v_{\text{B},x}) \\ \mu_y(z, \alpha, v_{\text{B},y}) \\ \mu_z(z, \alpha, v_{\text{B},z}) \end{bmatrix},
\end{equation}
where $\bm{\mu}$ denotes the posterior mean prediction of the trained Gaussian process model given the observed vector $\bm{z}$, and the matrix $\bm{B}_\text{z}$ defines the mapping between the subspace of the observed vector and the full state, i.e.,
\begin{equation}
  \bm{B}_\text{z} = \begin{bmatrix} \bm{0} \in \mathbb{R}^{3\times3} \\ \bm{0} \in \mathbb{R}^{4\times3} \\ \bm{I} \in \mathbb{R}^{3\times3} \end{bmatrix},
\end{equation}
where the identity matrix is denoted as $\bm{I}$.

The proposed NMPC formulation given the time-optimal trajectory $\bm{x}_\text{ref}$ is
\begin{equation}
  \label{eq:mpc_formulation}
  \begin{aligned}
    \min_{\bm{x}^*_{(\cdot)}, \bm{u}^*_{(\cdot)}} & \ \sum_{k=0}^{N-1}  (\| \bm{x}_{k} - \bm{x}_{\text{ref},k} \|_{\bm{Q}}^2+\| \bm{u}_{k} - \bm{u}_{\text{ref},k}\|_{\bm{R}}^2) \\ & \hspace{8mm} +\| \bm{x}_{N} - \bm{x}_{\text{ref},N} \|_{\bm{P}}^2  \\
    \text{s.t.} \ \
                                                  & \bm{x}_0 = \bm{x}(0),                                                                                                        \\
                                                  & \bm{x}_{k+1} = \bm{x}_{k} + \Delta t f_{\text{RK4}}(\bm{x}_{k}, \bm{u}_{k}) \ \forall  k\in  \{0,\ldots,N-1\},               \\
                                                  & \bm{x}_{\text{min}} \leq \bm{x}_{k} \leq \bm{x}_{\text{max}}\ \forall  k\in \{0,\ldots,N\},                                  \\
                                                  & \bm{u}_{\text{min}} \leq \bm{u}_{k} \leq \bm{u}_{\text{max}} \ \forall  \{0,\ldots,N-1\},                                    \\
  \end{aligned}
\end{equation}
where $\bm{Q}$ and $\bm{R}$ are positive-definite weighting matrices and $\bm{P}$ is a positive semi-definite weighting matrix, respectively. The function $f_\text{RK4}$ denotes the 4-th order explicit Runge-Kutta function of the augmented nonlinear dynamics $f_\text{augmented}(\bm{x},\bm{u})$. By solving the optimization problem for each time step, one can acquire the optimized control sequence $\bm{u}^*$ and the estimated state sequence $\bm{x}^*$ for the quadrotor, and then send the first optimized control input $\bm{u}^*_0$ to the flight controller. In this work, the proposed NMPC problem is implemented with the ACADOS library~\cite{VerschuerenFrisonKouzoupisFreyDuijkerenZanelliNovoselnikAlbinQuirynenDiehl2021}, and the optimal control input can be obtained within 0.01\,s. Furthermore, the GPR model is implemented with the GPU-accelerated libraries GPytorch~\cite{GardnerPleissBindelWeinbergerWilson2018} and Pytorch~\cite{PaszkeGrossMassaLererBradburyChananKilleenLinGimelsheinAntigaDesmaisonKopfYangDeVitoRaisonTejaniChilamkurthySteinerFangBaiChintala2019}, and each predicted result for three axes requires only 4~ms with an NVIDIA Quadro M1000 GPU.

\subsection{Experimental Setup}
\begin{figure}[htpb]
  \centering
  \includegraphics[scale=1]{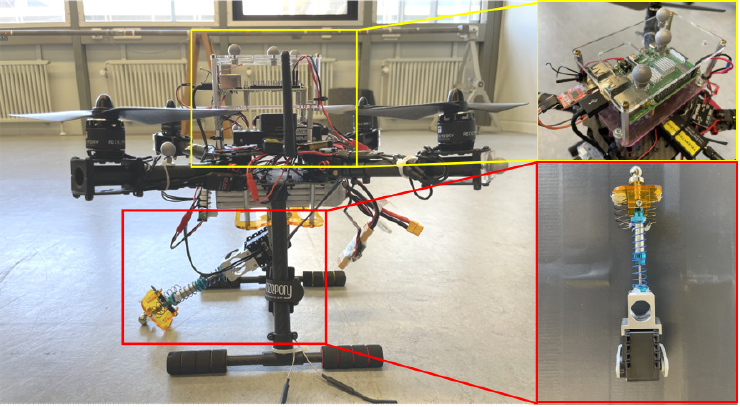}
  \caption{Employed aerial manipulator}
  \label{fig:hardware_aerial_manipulator}
\end{figure}

As illustrated in Fig.~\ref{fig:hardware_aerial_manipulator}, the proposed quadrotor with a 1-DoF manipulator consists of a quadrotor that with a motor-to-motor diagonal of 330~mm and a manipulator with a length of 182~mm. The latter is powered by a Dynamixel servo motor AX-12A. The flight control unit is the Pixhawk Cube Black with the open-source PX4 autopilot framework~\cite{MeierHoneggerPollefeys2015}. The onboard computer is a Raspberry Pi 4B+, which establishes the communication between the Pixhawk flight control unit and the laptop, which runs the NMPC controller and the proposed trajectory planner, via the local wireless network. The proposed control framework and the communication between devices are implemented in the ROS Noetic environment. Furthermore, the onboard manipulator is placed beneath the center of the quadrotor frame, where the end-effector is simplified with a permanent magnet and a shock absorber is used to attenuate impacts during handover.

\begin{figure}[htpb]
  \centering
  \includegraphics[scale=1]{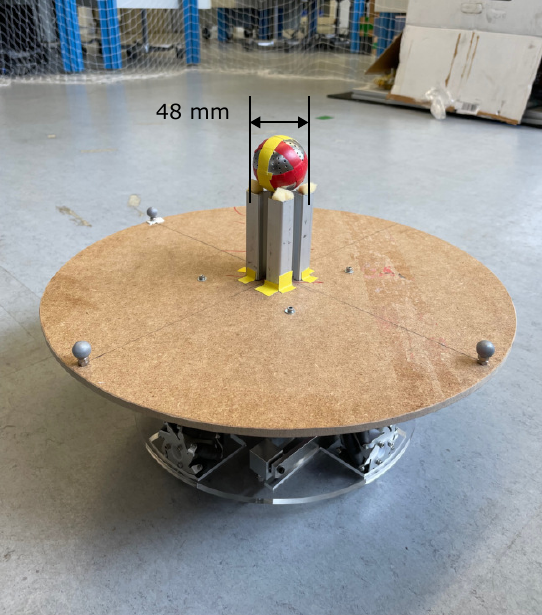}
  \caption{Mobile robot with the grasping object}
  \label{fig:mobile_robot}
\end{figure}

Furthermore, our omnidirectional mobile robot HERA~\cite{EschmannEbelEberhard2021} is holding a metal ball with a diameter of 4.8~cm that is constructed of iron, see Fig.~\ref{fig:mobile_robot}. The mobile robot is driven by a separate controller and follows the desired trajectory that, in this study, is assumed to be known a prior.

\subsection{Experimental Results}
For the handover hardware experiment, two distinct scenarios are prepared to verify the performance of the GPR-augmented NMPC framework given the planned time-optimal trajectory. Note that some physical and dynamical properties of the actual aerial manipulator may vary from those in the numerical simulation in Tab.~\ref{tab:numerical_simulation_parameters} due to the employed hardware and the scenario settings. Thus, the parameters pertaining to the real-world experiment are shown in Tab.~\ref{tab:hardware_parameters}. Parameters not contained in the latter are set to the values from Tab.~\ref{tab:numerical_simulation_parameters}.

\begin{table}[!t]
  \caption{Hardware Experiment Parameters}
  \centering
  \begin{tabular}{|m{3.5cm}|c|}
    \hline
    \centering
    parameters                                                                                          & value                                                  \\
    \hline
    \centering
    \vspace{1mm} quadrotor inertia \vspace{1mm}    $\bm{J}_\text{quadrotor}$                            & diag$\left(0.0158, 0.0154, 0.0195\right)$~[kg\,m$^2$]  \\
    \hline
    \centering
    \vspace{1mm}manipulator inertia \vspace{1mm}     $\bm{J}_\text{arm}$                                & diag$\left(0.0001, 0.0016, 0.00016\right)~$[kg\,m$^2$] \\
    \hline\centering
    \vspace{1mm}manipulator  offset $\bm{\ell}_\text{offset,B}\vspace{1mm}$                             & $\left[0 \ 0 \ -0.107\right]^\top$~[m]                 \\
    \hline\centering
    \vspace{1mm}max. allowed heading difference by handover $c_\text{limit, heading}$      \vspace{1mm} & 0.01~[rad]                                             \\
    \hline\centering
    \vspace{1mm}max./min. velocity               \vspace{1mm}                                           & $\left[\pm0.3 \ \pm0.3 \ \pm0.15\right]^\top$~[m/s]    \\
    \hline\centering
    \vspace{1mm}max./min. angular velocity           \vspace{1mm}                                       & $\left[\pm0.2 \ \pm0.2 \ \pm0.1 \right]^\top$~[rad/s]  \\
    \hline\centering
    \vspace{1mm}max./min. manipulator rotation velocity                           \vspace{1mm}          & $\pm \pi/5$~[rad/s]                                    \\
    \hline
  \end{tabular}
  \label{tab:hardware_parameters}
\end{table}

\begin{figure}[htpb]
  \centering
  \includegraphics[scale=1]{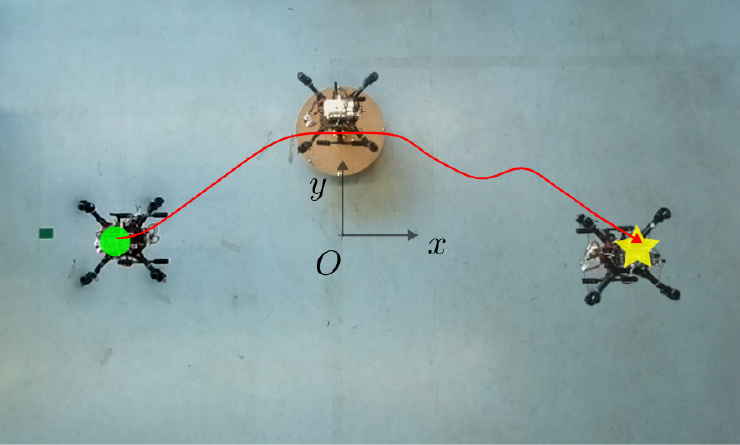}
  \caption{Static object scenario. The red line indicates the desired trajectory projected to the $x$-$y$-plane. The green point denotes the desired start point, and the yellow star denotes the desired end point, respectively.}
  \label{fig:fp_exp_xy_sc}
\end{figure}

\subsubsection{Static Object Grasping}
\begin{figure}[htpb]
  \centering
  \includegraphics[scale=1]{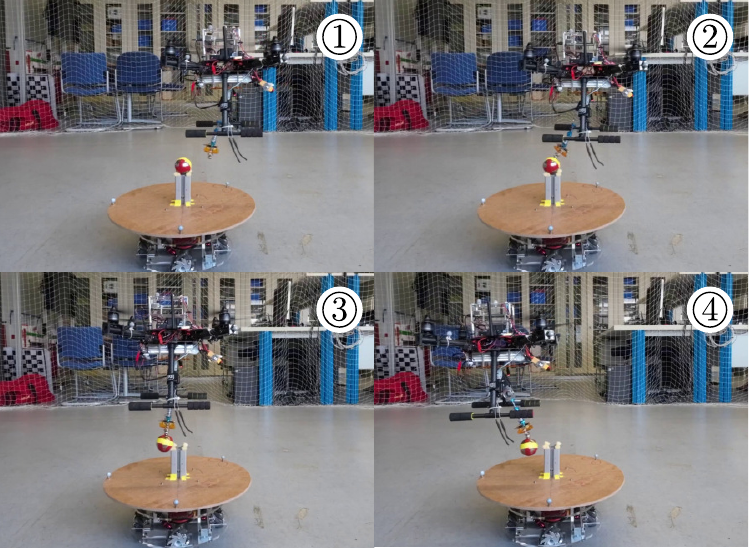}
  \caption{Photographs of the hardware experiment when grasping a static object}
  \label{fig:fp_exp_sc}
\end{figure}

\begin{figure}[htpb]
  \centering
  \subfloat[poses of the aerial manipulator]{\includegraphics[scale=0.7]{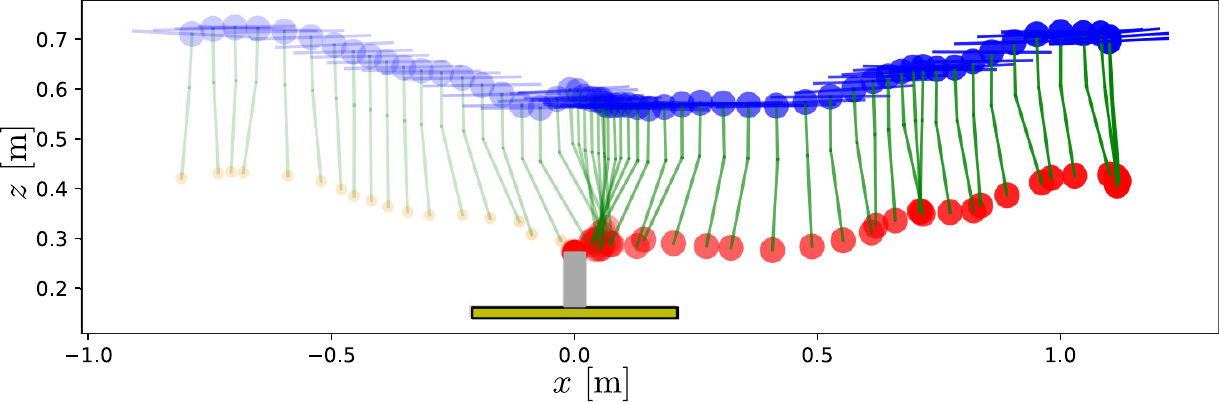}%
    \label{fig:fp_xz}}
  \hfil
  \subfloat[trace of the aerial manipulator's end-effector]{\includegraphics[scale=0.7]{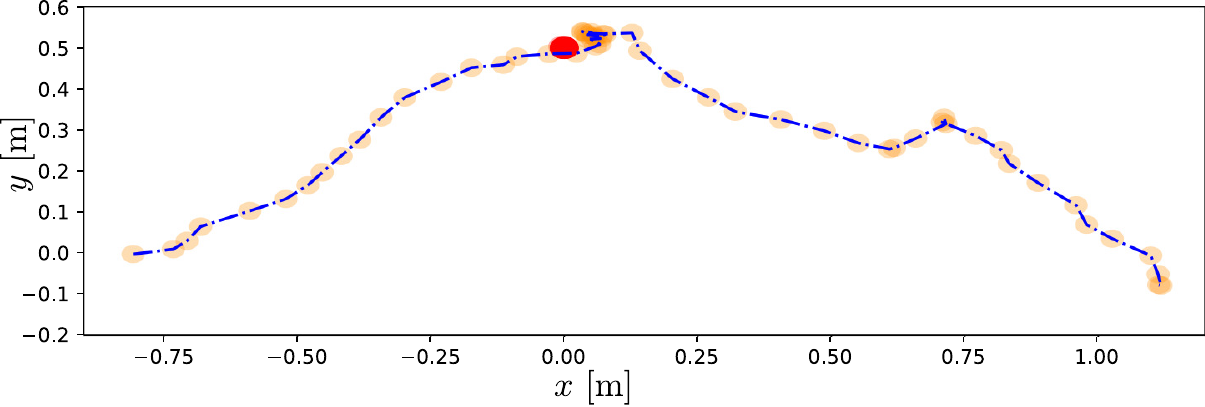}%
    \label{fig:fp_mani_xy}}
  \hfil
  \subfloat[distance between the aerial manipulator's end-effector and the desired handover position]{\includegraphics[scale=0.7]{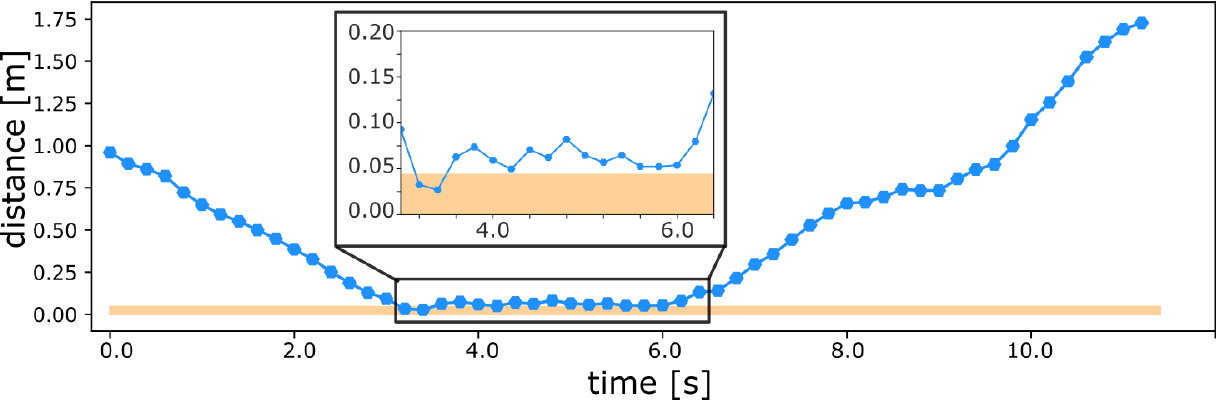}%
    \label{fig:fp_ee_distance}}
  \caption{Fixed object grasping results}
  \label{fig:fp_results}
\end{figure}

In the first scenario, the mobile robot is placed on a fixed position during the mission, and the planned trajectory based on the proposed trajectory planning framework is illustrated in Fig.~\ref{fig:fp_exp_xy_sc}. Based on the proposed trajectory planning algorithm and the provided control framework, our aerial manipulator has successfully grasped the static object, and four time steps surrounding the moment of grasping are demonstrated in Fig.~\ref{fig:fp_exp_sc}. Furthermore, Fig.~\ref{fig:fp_xz} illustrates the poses of the aerial manipulator throughout the grasping process in accordance with the indoor localization system in our laboratory based on OptiTrack, while the opacity of each depicted pose increases as time passes.

To guarantee a steady handover flight, it should be noted that in this experiment, as opposed to the numerical simulation setup, the initial value of $\kappa_\text{init}$ is set to 13, so that the contact period is extended. The position of the end-effector in the $x$-$y$-plane is shown in Fig.~\ref{fig:fp_mani_xy}, where the position of the target is marked with red circles, and the region in which the permanent magnet may grasp the object is marked with orange circles, and the trace of the end-effector is shown by a dash-dotted line. Moreover, the distance changes between the end-effector and desired handover position are shown in Fig.~\ref{fig:fp_ee_distance}, where the acceptable contact distance is marked with orange. Here, the desired handover position refers to the location of the grasping object when it is placed above the mobile robot as shown in Fig.~\ref{fig:mobile_robot}.

Remark that, since the permanent magnet is not controllable, once the end-effector has contact with the target ball, the ball remains attached to the end-effector. However, the disturbances caused by the grasped target are not considered in our proposed controller since this work's focus is the grasping process. As a direct consequence of this, the tracking performance once the grasping is successfully completed is negatively impacted, and the tracking inaccuracy is subsequently increased.

\subsubsection{Circular Trajectories with Different $\kappa_\text{init}$ }
\begin{figure}[htpb]
  \centering
  \includegraphics[scale=1]{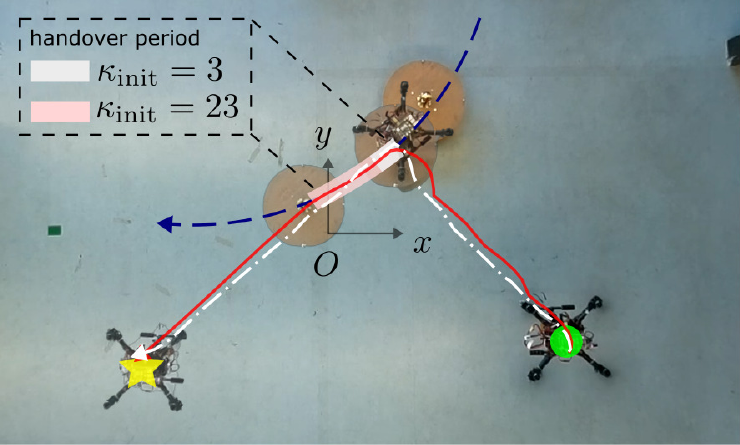}
  \caption{Moving object scenario. Two planned time-optimal trajectories are illustrated with different $\kappa_\text{init}$.}
  \label{fig:circle_exp_xy_sc}
\end{figure}

In the second experiment, the proposed trajectory planning framework and the designed control framework shall handle a dynamic scenario in which the mobile robot is traveling along a circular trajectory. The latter is illustrated in Fig.~\ref{fig:circle_exp_xy_sc} with a line dashed in blue. Furthermore, to show the impact of varying initial values of $\kappa_\text{init}$ on the planned trajectory, two different $\kappa_\text{init}$ are tested, and the planned time-optimal trajectories for hand-over are shown in Fig.~\ref{fig:circle_exp_xy_sc}, where the white dash-dotted line represents the results from $\kappa_\text{init}=3$ and the red line denotes the results from $\kappa_\text{init}=23$, respectively. The planned results show that for a smaller $\kappa_\text{init}$, the generated trajectory has a considerably shorter contact duration. On the contrary, the larger $\kappa_\text{init}$ lets the aerial manipulator stay in close company with the object for a longer time frame, and the generated trajectory closely follows the trajectory of the mobile robot during this time frame.

\begin{figure}[htpb]
  \centering
  \includegraphics[scale=1]{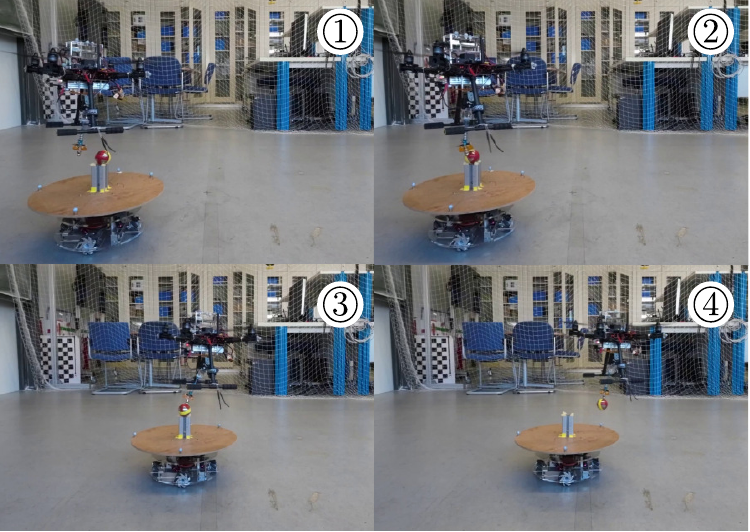}
  \caption{Photographs of the hardware experiment when grasping a moving object ($\kappa_\text{init}=23$)}
  \label{fig:circle_exp_sc}
\end{figure}

\begin{figure}[htpb]
  \centering
  \subfloat[poses of the aerial manipulator]{\includegraphics[scale=0.7]{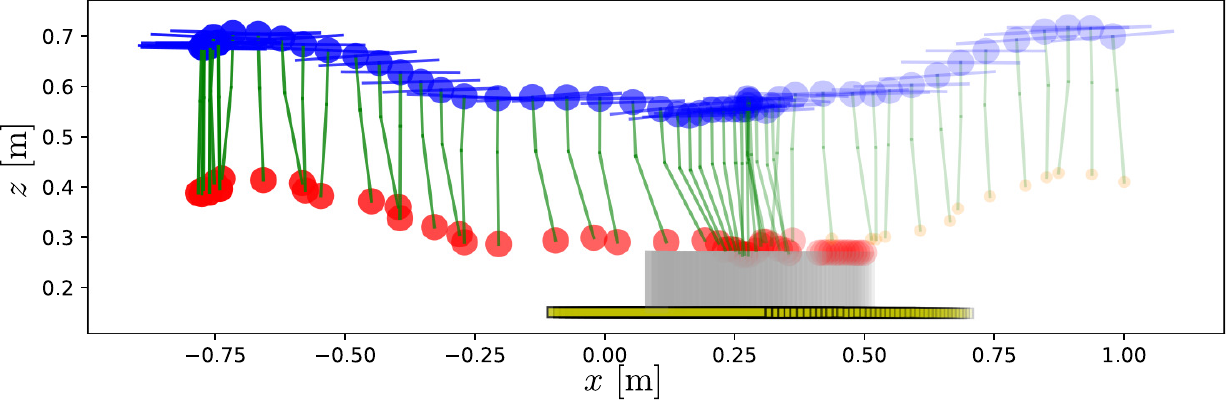}%
    \label{fig:circle_xz_k3}}
  \hfil
  \subfloat[trace of the aerial manipulator's end-effector]{\includegraphics[scale=0.7]{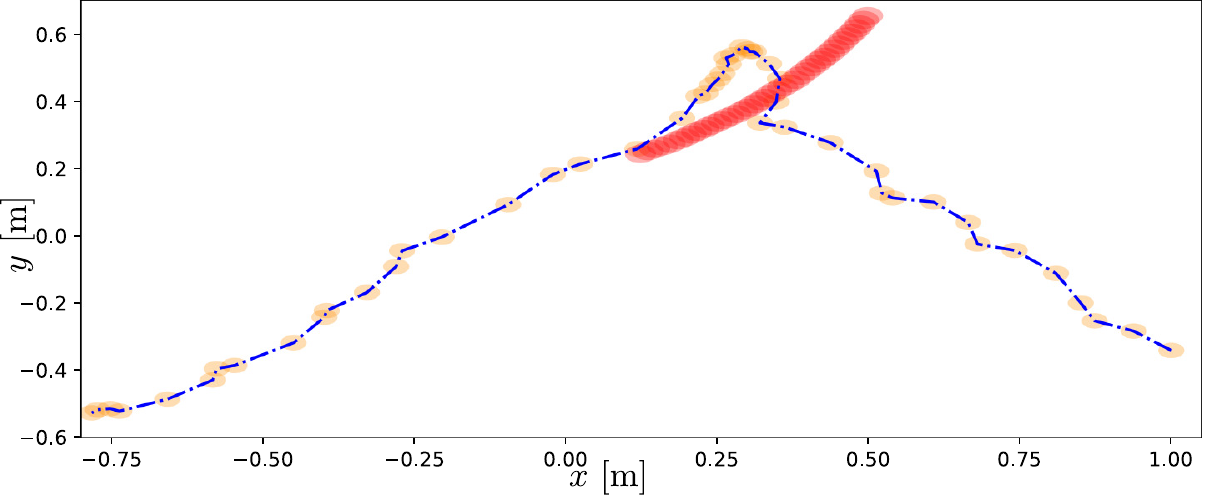}%
    \label{fig:circle_mani_xy_k3}}
  \hfil
  \subfloat[distance between the aerial manipulator's end-effector and the desired handover position]{\includegraphics[scale=0.7]{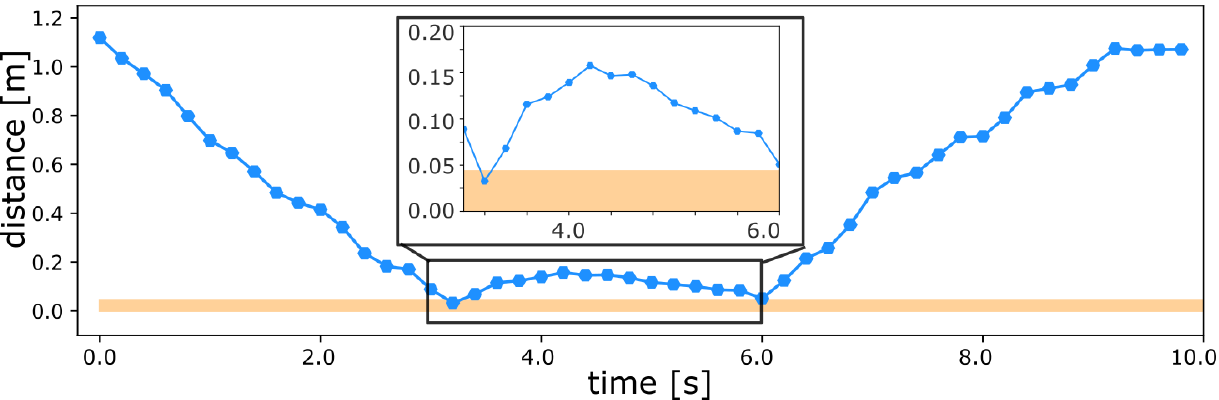}%
    \label{fig:circle_ee_distance_k3}}
  \caption{Moving object grasping results ($\kappa_\text{init}=3$)}
  \label{fig:circle_results_k3}
\end{figure}

\begin{figure}[htpb]
  \centering
  \subfloat[poses of the aerial manipulator]{\includegraphics[scale=0.7]{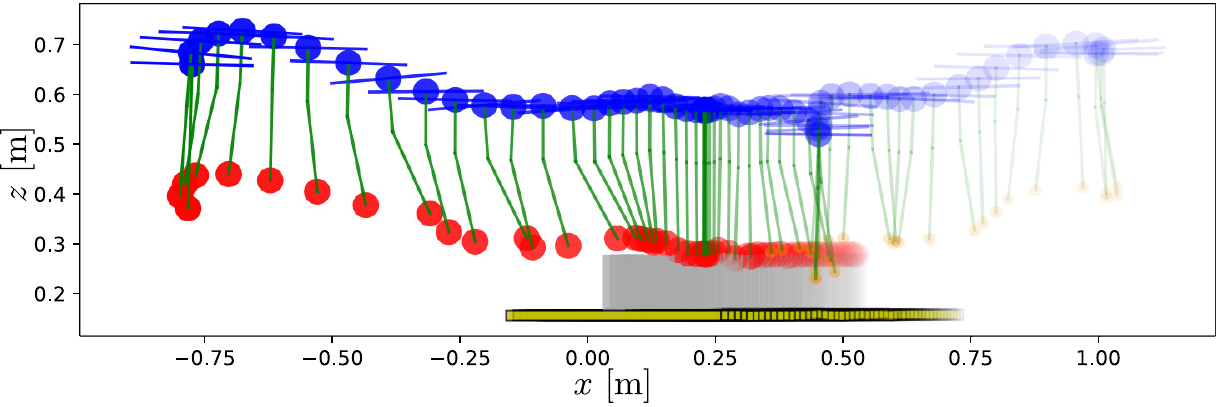}%
    \label{fig:circle_xz_k23}}
  \hfil
  \subfloat[trace of the aerial manipulator's end-effector]{\includegraphics[scale=0.7]{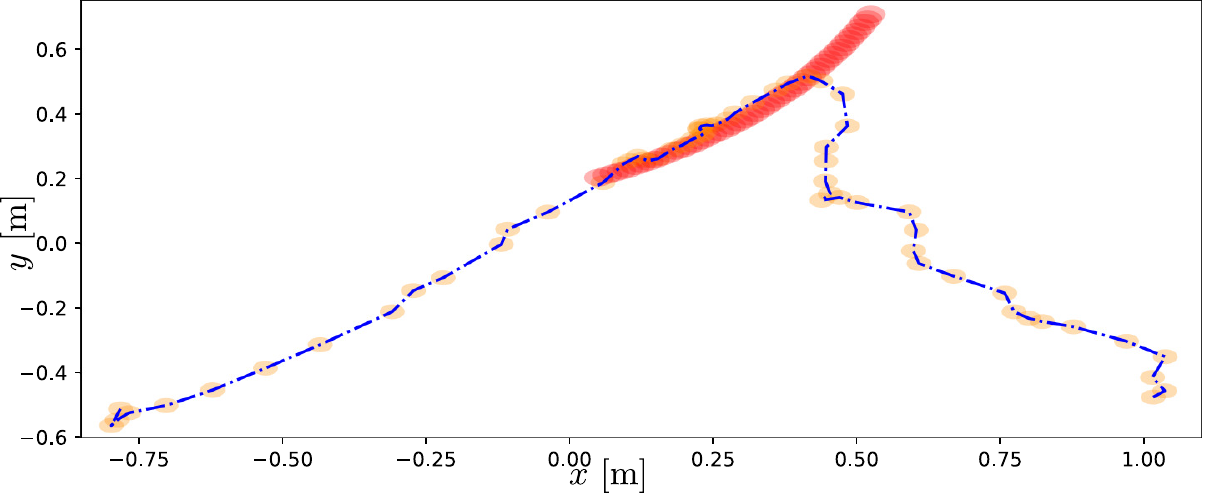}%
    \label{fig:circle_mani_xy_k23}}
  \hfil
  \subfloat[distance between the aerial manipulator's end-effector and the desired handover position]{\includegraphics[scale=0.7]{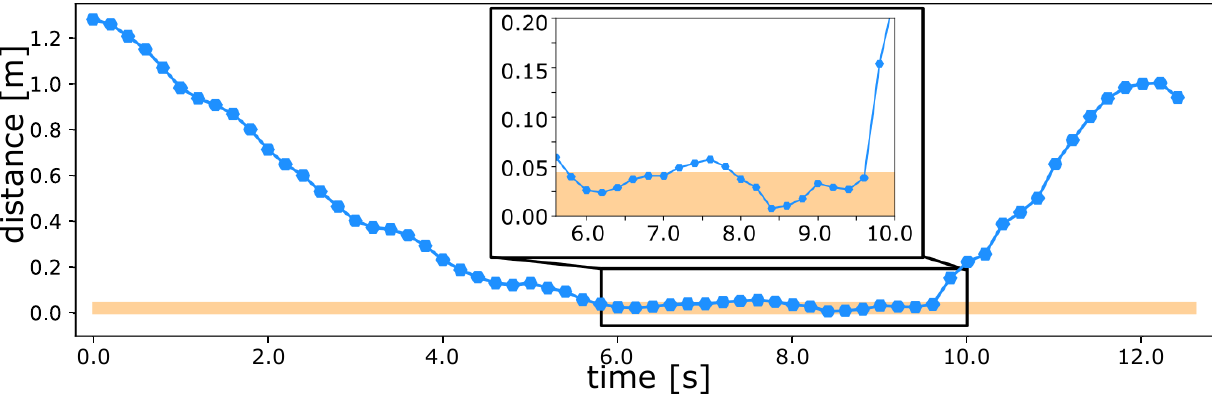}%
    \label{fig:circle_ee_distance_k23}}
  \caption{Moving object grasping results ($\kappa_\text{init}=23$)}
  \label{fig:circle_results_k23}
\end{figure}

The hardware experiment in our laboratory is depicted in Fig.~\ref{fig:circle_exp_sc}, and the real-flight performance given the planned time-optimal trajectories with the corresponding $\kappa_\text{init}$ is shown in Figs.~\ref{fig:circle_results_k3} and~\ref{fig:circle_results_k23}, respectively. In both experiments, the planned time-optimal trajectories can be executed with our aerial manipulator based on the proposed NMPC controller, and the moving target ball is grasped successfully. Yet, compared to performance with a smaller $\kappa_\text{init}$, the aerial manipulator has a significantly longer handover period given the trajectory planned with $\kappa_\text{init}=23$ as shown in Figs.~\ref{fig:circle_xz_k3} and~\ref{fig:circle_xz_k23}, which corresponds to the planned trajectories depicted in Fig.~\ref{fig:circle_exp_xy_sc}. Furthermore, the total operation time is also impacted by different $\kappa_\text{init}$, and the calculated optimal entire travel time $t_N$ is slightly increased due to the larger $\kappa_\text{init}$ as indicated in Figs.~\ref{fig:circle_ee_distance_k3} and~\ref{fig:circle_ee_distance_k23}.


\section{Conclusion and Future Work}
In this work, we propose an optimization framework based on the discrete mechanics and complementarity constraints (DMCC) to design an admissible time-optimal trajectory for the aerial manipulator while accomplishing a handover task, where the timing of the handover is determined automatically, while conforming to the studied aerial manipulator's dynamics. This framework is initially verified in several numerical simulations, including aerial racing and hand-over scenarios. In real-world experiments, the proposed DMCC framework has proven to generate appropriate trajectories for our customized aerial manipulator to grasp a small object from a mobile robot.

In future work, instead of treating the trajectory prediction of the observed target as an individual task, one may incorporate the target trajectory prediction into the optimization and address the entire issue simultaneously. Moreover, future work may strive to account for the influence of the carried object on the dynamics in the controller after having picked up the object.

\section*{Acknowledgment}
This work was supported by the Deutsche Forschungsgemeinschaft
(DFG, German Research Foundation) under Grant 433183605 and through
Germany's Excellence Strategy (Project PN4-4 Theoretical Guarantees
for Predictive Control in Adaptive Multi-Agent Scenarios) under Grant
EXC 2075-390740016. This research also benefited from the support by the China Scholarship Council (CSC, No.~201808080061) for Wei Luo.

\ifCLASSOPTIONcaptionsoff
  \newpage
\fi



\bibliographystyle{IEEEtran}
\bibliography{bib_files/tro_2022}
%



%

\begin{IEEEbiography}[{\includegraphics[width=1in,height=1.25in,clip,keepaspectratio]{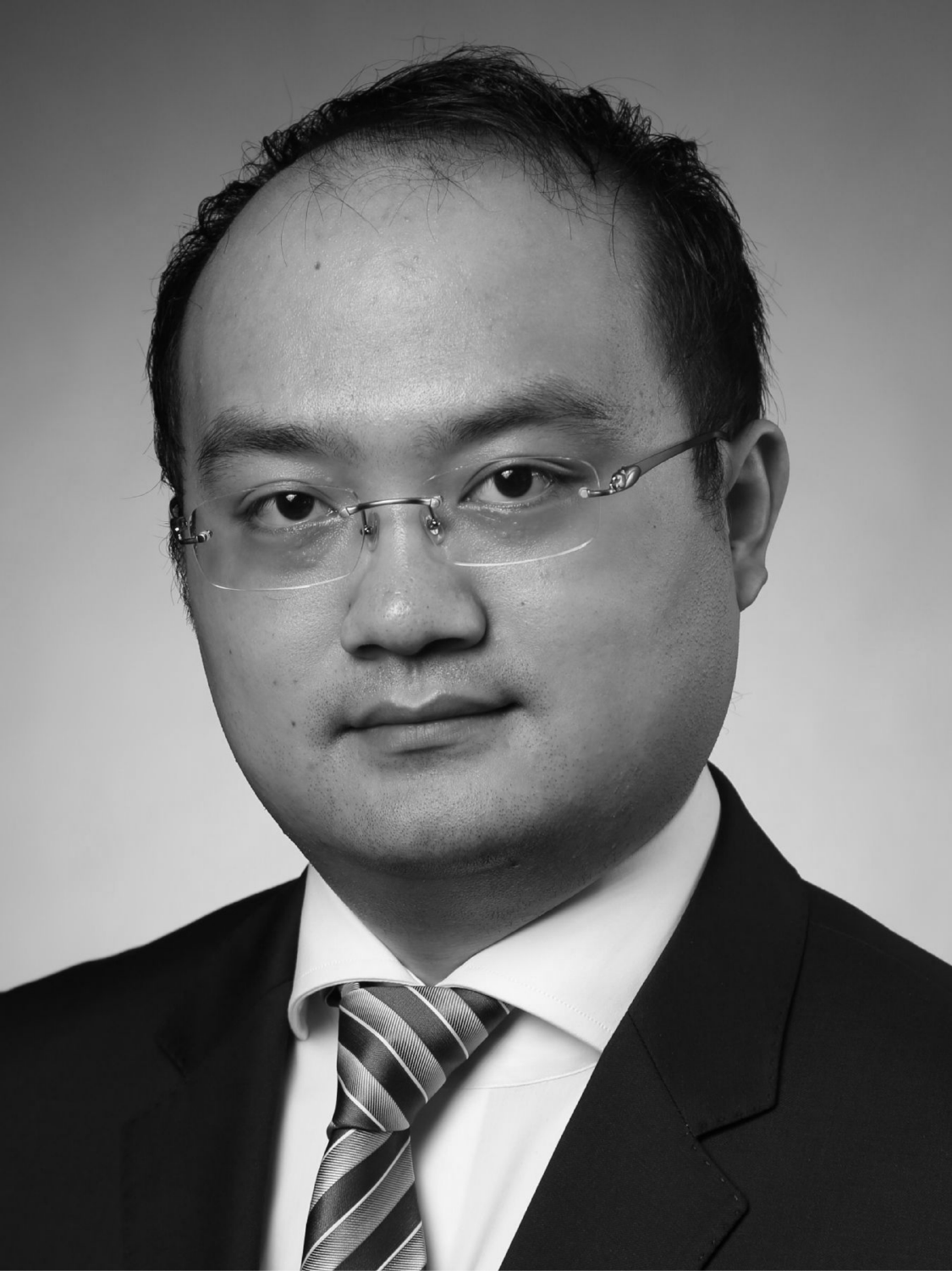}}]{Wei Luo}
  received the B.Sc.~degree in 2013 and M.Sc.~degree in 2016 in mechanical engineering from the University of Stuttgart, Stuttgart, Germany. There he is currently pursuing doctoral studies with the Institute of Engineering and Computational Mechanics, University of Stuttgart. He is a member of the research staff there. His research interests include robotics and multibody systems, especially the control and the application to heterogeneous swarm robot system.
\end{IEEEbiography}

\begin{IEEEbiography}[{\includegraphics[width=1in,height=1.25in,clip,keepaspectratio]{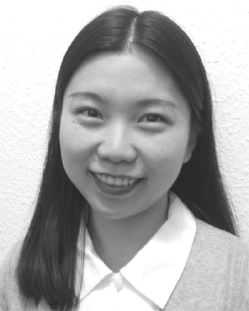}}]{Jingshan Chen}
  received the B.Sc.~degree in mechanical engineering and automation from China University of Mining and Technology-Beijing in 2016 and M.Sc.~degree in mechatronics from the University of Stuttgart, Germany in 2021. She is currently pursuing doctoral studies with the Institute of Engineering and Computational Mechanics, University of Stuttgart and is a member of the research staff there. Her research interests include dynamics of multibody systems and control theory, especially in its applications to robotics.
\end{IEEEbiography}

\begin{IEEEbiography}[{\includegraphics[width=1in,height=1.25in,clip,keepaspectratio]{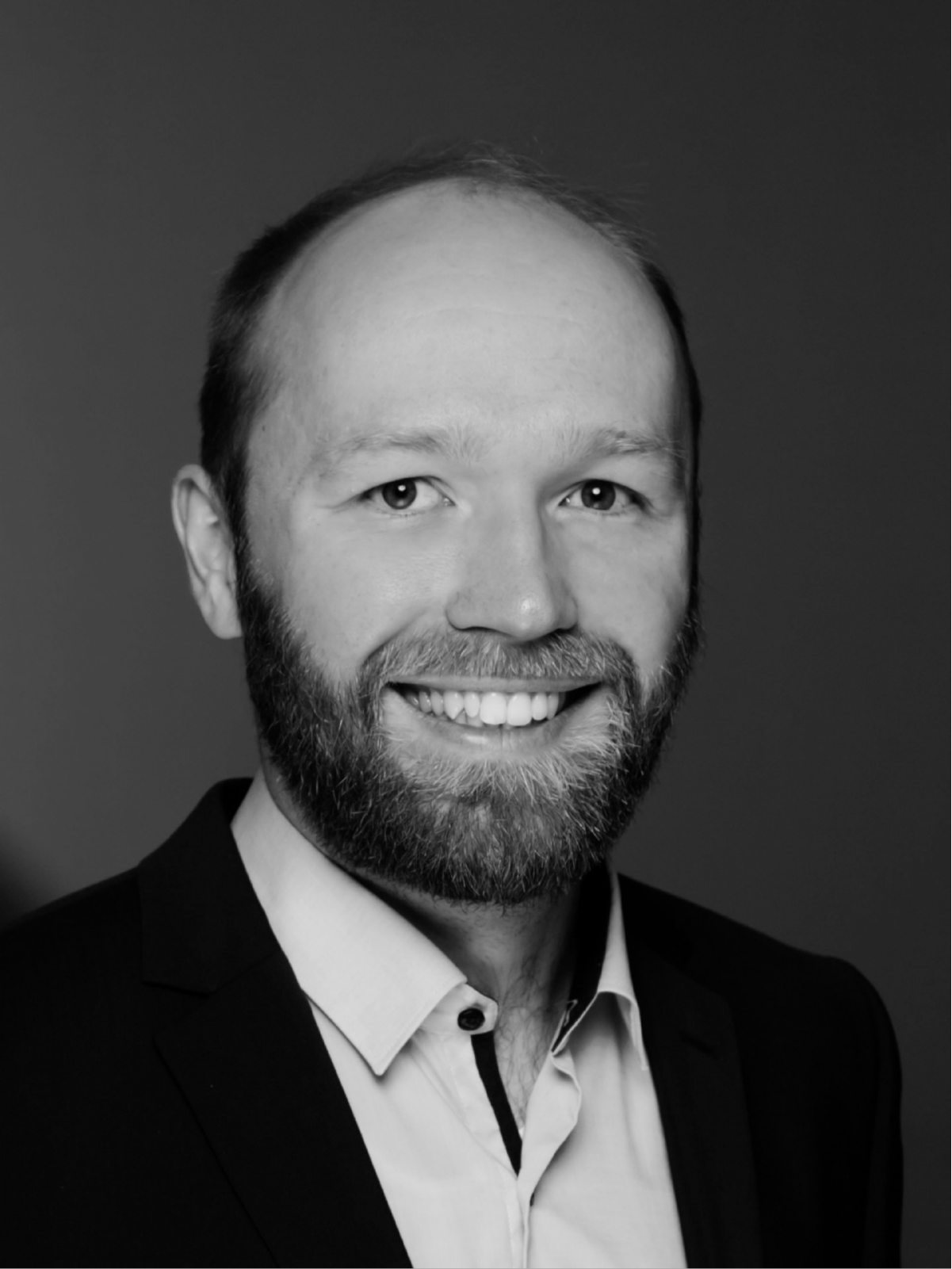}}]{Henrik Ebel}
  received his B.Sc. and M.Sc. degrees in Simulation Technology from the University of Stuttgart, Germany, in 2014 and 2016, and his doctoral degree in 2021. He is currently a postdoctoral researcher and a member of the research staff at the Institute of Engineering and Computational Mechanics at the University of Stuttgart. His research interests include multibody system dynamics and control theory, especially in its applications to mechanical systems and in the field of robotics. Of particular interest are the cooperation of multiple robotic agents, as well as optimization-based control schemes.
\end{IEEEbiography}

\begin{IEEEbiography}[{\includegraphics[width=1in,height=1.25in,clip,keepaspectratio]{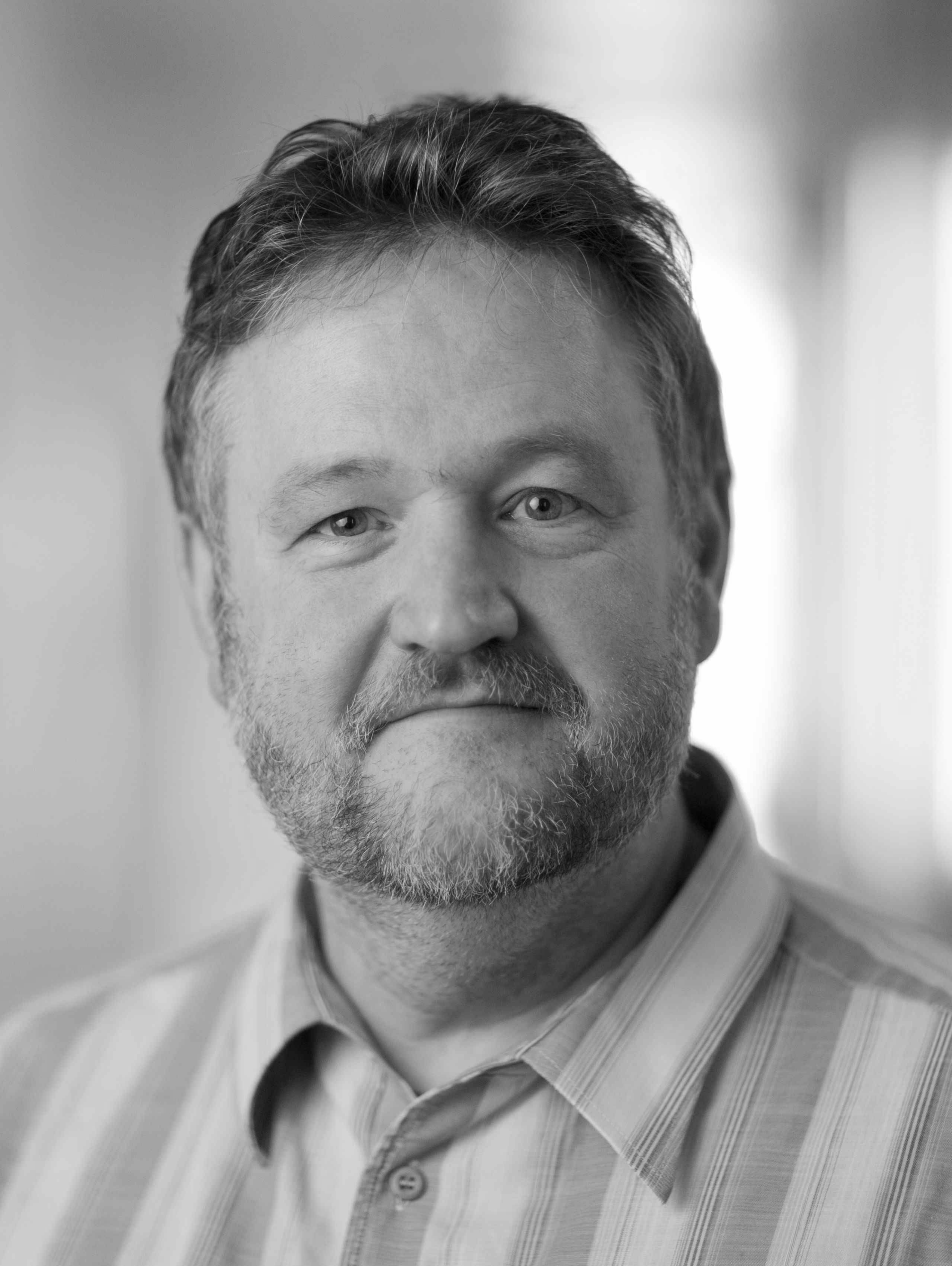}}]{Peter Eberhard}
  is full professor in mechanics/dynamics and since 2002 director of the Institute of Engineering and Computational Mechanics (ITM) at the University of Stuttgart, Germany. He was Treasurer and Bureau member of IUTAM, the International Union of Theoretical and Applied Mechanics, and served before in many national and international organizations, e.g., as Chairman of the IMSD (International Association for Multibody System Dynamics) or DEKOMECH (German Committee for Mechanics).
\end{IEEEbiography}







\end{document}